\pdfoutput=1

\documentclass[11pt]{article}

\usepackage{acl}

\usepackage{times}
\usepackage{latexsym}

\usepackage[T1]{fontenc}

\usepackage[utf8]{inputenc}

\usepackage{microtype}

\usepackage{inconsolata}

\usepackage{booktabs}
\usepackage{tabularx}
\usepackage{graphicx}
\usepackage[most]{tcolorbox}
\usepackage{soul}
\usepackage{multirow}
\usepackage{xspace}
\usepackage{subcaption}
\usepackage{enumitem}
\usepackage{cleveref}
\crefname{section}{§}{§§}
\Crefname{section}{§}{§§}

\newcommand{\papercomment}[3]{{\textcolor{#3}{[#1 #2]}}}
\newcommand{\marker}[1]{\textbf{#1:} }
\newif\ifcomments
\ifcomments
    \newcommand{\sunipa}[1]{\papercomment{\marker{sunipa}}{#1}{cyan}}
    \newcommand{\sameer}[1]{\papercomment{\marker{sameer}}{#1}{purple}}
    \newcommand{\tamanna}[1]{\papercomment{\marker{tamanna}}{#1}{magenta}}
    \newcommand{\anthony}[1]{\papercomment{\marker{anthony}}{#1}{violet}}
\else
    \newcommand{\sunipa}[1]{}%
    \newcommand{\sameer}[1]{}%
    \newcommand{\tamanna}[1]{}%
    \newcommand{\anthony}[1]{}%
\fi

\definecolor{mintgreen}{HTML}{98FF98}
\definecolor{softrose}{HTML}{FFB6C1}
\definecolor{rose}{HTML}{D71D57}
\definecolor{emeraldgreen}{HTML}{009688}
\definecolor{slateblue}{HTML}{6A5ACD}
\definecolor{mediumorchid}{HTML}{BA55D3}
\definecolor{fallred}{HTML}{9A031E}
\definecolor{fallteal}{HTML}{0f4c5c}
\definecolor{fallgreen}{HTML}{606c38}
\definecolor{cerisepink}{HTML}{E91E63}
\definecolor{lavender}{HTML}{755D9A}
\definecolor{pastelyellow}{HTML}{FFF9B0}
\definecolor{pastelblue}{HTML}{AEDFF7}

\newcommand{\misg}{{\color{fallred}\texttt{Misgendering}}\xspace}
\newcommand{\nomisg}{{\color{fallgreen}\texttt{No Misgendering}}\xspace}

\newcommand{\dataset}
{{\color{lavender}\textsc{MisgenderMender}}\xspace}

\newtcbox{\mybox}[1]{
    on line, %
    arc=4pt, 
    auto outer arc,
    outer arc=0pt,
    colback=#1,
    colframe=#1,
    boxrule=0pt,
    boxsep=0pt,
    left=0pt, 
    right=0pt, 
    top=1pt, 
    bottom=1pt 
}
\newcommand{\detect}{\mybox{pastelyellow}{Detect-Only} }
\newcommand{\edit}{\mybox{pastelblue}{Detect+Correct} }

\title{\dataset: \\ A Community-Informed Approach to Interventions for Misgendering}

\author{Tamanna Hossain \\
  University of California, Irvine\\
 \href{mailto:tthossai@uci.edu}{\texttt{tthossai@uci.edu}}   \\\And
  Sunipa Dev \\
  Google Research \\
 \href{mailto:sunipadev@google.com}{\texttt{sunipadev@google.com}} \\\And
  Sameer Singh\\
  University of California, Irvine \\
  \href{mailto:sameer@uci.edu}{\texttt{sameer@uci.edu}}
  }

\begin{document}
\maketitle

\begin{abstract}
\textcolor{red}{\textit{Content Warning:} This paper contains examples of misgendering and erasure that could be offensive and potentially triggering.}

Misgendering, the act of incorrectly addressing someone's gender, inflicts serious harm and is pervasive in everyday technologies, yet there is a notable lack of research to combat it.
We are the first to address this lack of research into interventions for misgendering by conducting a survey of gender-diverse individuals in the US to understand perspectives about automated interventions for text-based misgendering.
Based on survey insights on the prevalence of misgendering, desired solutions, and associated concerns, we introduce a misgendering interventions task and evaluation dataset, \dataset.%
We define the task with two sub-tasks: (i) detecting misgendering, followed by (ii) correcting misgendering where misgendering is present in domains where editing is appropriate.
\dataset comprises 3790 instances of social media content and LLM-generations about non-cisgender public figures, annotated for the presence of misgendering, with additional annotations for correcting misgendering in LLM-generated text.
Using this dataset, we set initial benchmarks by evaluating existing NLP systems and highlighting challenges for future models to address.
We release the full dataset, code, and demo at \url{https://tamannahossainkay.github.io/misgendermender/}.
\end{abstract}

\begin{figure}[!tb]
\small
\begin{tabularx}{\columnwidth}{@{}lX@{}}
\toprule
\multicolumn{2}{c}{Linguistic Gender Profile}                                                                                                    \\ \midrule
\textbf{Name:}                           & Elliot Page                                                           \\
\textbf{Gender identity:}                & Trans man, Non-binary                                              \\
\textbf{Pronouns:}                       & he/him/his/his/himself, they/them/their/theirs/themselves        \\
\textbf{Gendered Terms:} & masculine, neutral\\
\textbf{Deadname:}                       & Ellen Grace Philpotts-Page                                            \\
\bottomrule
\end{tabularx}
\begin{tabularx}{\columnwidth}{@{}X@{}}
\toprule
\multicolumn{1}{c}{Annotated Content}\\
\midrule
\detect \\
\textbf{X Post:} John Wayne was a man and Elliot Page is a woman… \\
\textbf{Detect Label:} \misg \\
\addlinespace
\textbf{X Post:} ..."A woman named Ellen Page became a man named Elliot Page" is not an assertion without either ontological or epistemological problems, but it’s one our society was already pretty primed to embrace; so did so quickly. \\
\textbf{Detect Label:} \nomisg \\
\addlinespace
\midrule
\edit \\
\textbf{LLM-generation:}  Ellen Grace credits her mother with her success, and she is eternally grateful for her love and support. \\
\textbf{Detect Label:} \misg \\
\textbf{Corrected:} \textcolor{fallred}{\st{Ellen }}$ \rightarrow $\textcolor{fallgreen}{ Elliot} credits  \textcolor{fallred}{\st{her }}$ \rightarrow $\textcolor{fallgreen}{ his} mother with  \textcolor{fallred}{\st{her }}$ \rightarrow $\textcolor{fallgreen}{ his} success, and  \textcolor{fallred}{\st{she }}$ \rightarrow $\textcolor{fallgreen}{ he} is eternally grateful for her love and support.\\
\bottomrule
\end{tabularx}
\caption{\dataset examples consisting of a gender linguistic profile and corresponding annotated content for detecting and correcting misgendering.} 
\label{data_example}
\end{figure}

\section{Introduction}
Misgendering is the act of referring to someone using a word, e.g. a pronoun or title, that does not correctly reflect the gender with which they identify \cite{oed_misgender}.
While there is growing awareness about the adverse impacts of misgendering on peoples' lives \cite{dev2021harms}, there is insufficient scholarship or resources that identify and attempt to mitigate misgendering in these various daily use platforms and technologies.

Efforts to measure and mitigate gender bias in natural language processing primarily focus on cisgender and binary gender categories \citep{guo-etal-2022-auto, choubey-etal-2021-gfst}.
Few efforts to address non-traditional gender categories have evaluated LLMs' abilities to use non-binary pronouns \cite{hossain-etal-2023-misgendered}, coreference resolution using neo-pronouns \cite{cao-daume-iii-2020-toward}, and representational biases in word embeddings \cite{dev2021harms}.
Furthermore, even though misgendering is both a factual inaccuracy and a toxic act of identity erasure, research on factuality and toxicity has largely ignored it \cite{gao-emami-2023-turing, Lees2022ANG}. 

Our contribution is two-fold: (i) we conduct a community survey to understand opinions about automated interventions for text-based misgendering, and (ii) based on the survey, we define a task and evaluation dataset for addressing misgendering in text-based content. 
Our survey of gender-diverse\footnote{Individuals who self-identify as non-cisgender or have changed their gender terminology at some point in their lives} individuals revealed a prevalent issue of misgendering, especially on social media, but also in other areas like AI-generated content, news articles, and academic journals (\cref{survey}).
While there was a general preference for automatic detection of misgendering across domains, opinions diverged on measures such as correcting or hiding misgendered content (\cref{survey_pref}). 
Participants were more receptive to the idea of auto-correction in AI-generated content than social media, citing concerns over limiting freedom of speech and creating a false sense of allyship. 
Importantly, there were significant apprehensions regarding the implementation of any automated systems to address misgendering, encompassing issues like the fundamental infeasibility of these systems, privacy, the risk of profiling or targeting based on gender linguistic preferences databases, and doubts about the current capabilities of NLP systems to perform interventions accurately (\cref{survey_concerns}).

Based on the opinions and concerns expressed by participants in our survey, we defined a task for misgendering interventions and constructed a corresponding evaluation dataset, \dataset (\cref{dataset}).
We define the interventions for misgendering task as two sub-tasks: (i) detecting misgendering, followed by (ii) correcting misgendering where misgendering is present, in domains where editing is appropriate (\cref{prob_setup}). 
Social media (X and YouTube) were picked as a \detect domain and LLM-generations as a \edit domain.
Text from each of these sources was collected regarding 30 non-cisgender public figures whose gender identities and gender terminology preferences are publicly available (\cref{data}).
A total of 3790 instances are human annotated for the misgendering interventions task (\cref{ann}).
See Figure \ref{data_example} for examples from \dataset dataset.

We evaluated current NLP systems using \dataset, setting initial benchmarks and pinpointing areas for future work. 
For the detection sub-task, we prompted language models using similar instructions to those given to human annotators, including providing the gender linguistic profile of the relevant individual.
We also used toxicity detection and rule-based baselines (\cref{detect}).
GPT-4 achieved the highest F1-score across domains, but there is still much room for improvement (\textit{X posts}: 62.6, \textit{YouTube Comments}: 85.3, \textit{LLM-generations}: 55.9).
There were errors associated with coreference resolution, understanding questions, temporal relationships, quotations, and authorship recognition.
For the second sub-task of correcting misgendering, we used a rule-based editor and prompting of GPT-4 (\cref{edit}).
Human evaluation of edits showed GPT-4 corrected misgendering in 97\% of edits while making unnecessary edits in only 4.6\% of cases.
While this is promising, further work is still needed since these edits were largely single-sentence and context-free.
To facilitate this, we release the full dataset, code, and demo of our work at \url{https://tamannahossainkay.github.io/misgendermender/}.

\begin{figure*}[!ht]
    \centering
    \small
        \includegraphics[width=\textwidth]{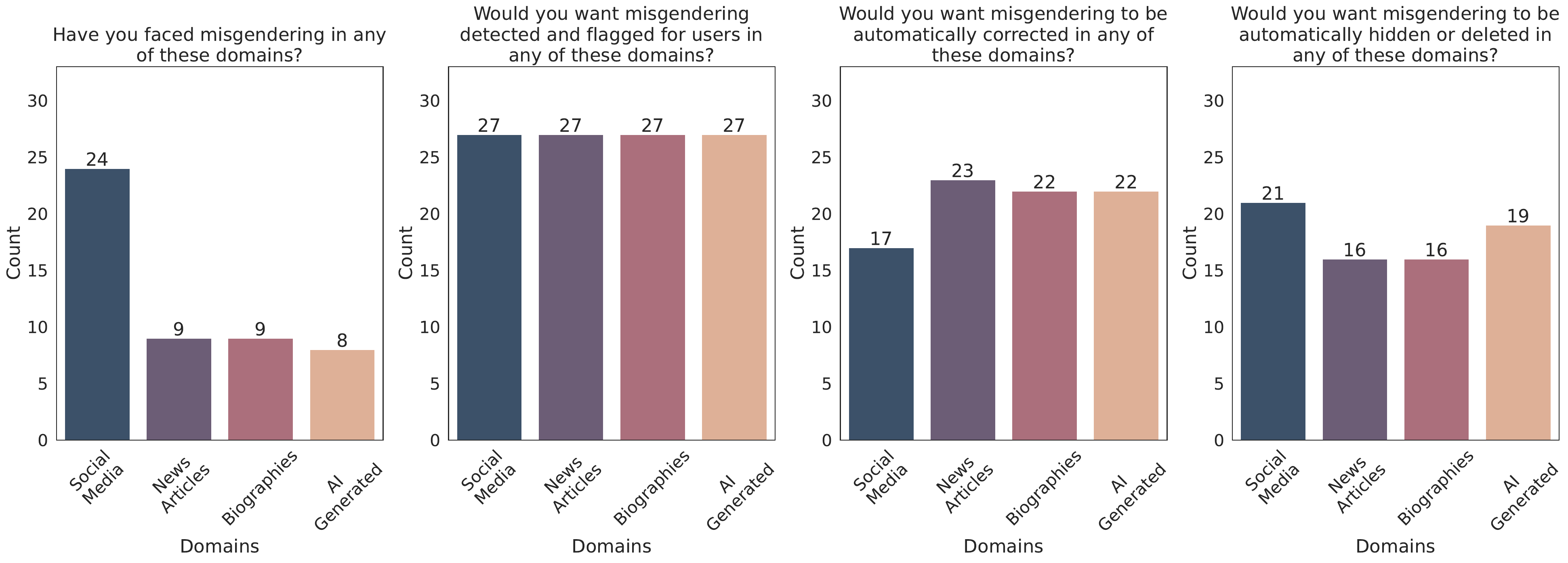}    
        \caption{\textbf{Survey responses} Count of participants (out of 33) reporting experiences with misgendering and expressing a desire for detection, correction, or hiding of misgendering across various domains.}
    \label{fig:survey}
\end{figure*}

\section{Survey on Interventions for Misgendering}
\label{survey}
Automated systems to prevent misgendering lack existing research. 
In order to define a task and develop an evaluation dataset rooted in community perspectives, we first survey gender-diverse individuals on their views regarding automated interventions for misgendering.

\paragraph{Methodology} The survey is anonymous and is conducted using Google Forms. 
We do not collect any data which could personally identify respondents. 
We reached out to participants through Queer in AI, International Society of Non-binary Scientists (ISBNS), and social media. 
All participants were adults (18 years or older) living in the US, who either identified as non-cisgender or had changed their gender terminology at some point in their lives.
The survey consists of four sections, which solicit participants' demographic data, experiences with misgendering, preferences for misgendering interventions, concerns regarding automated intervention systems, and miscellaneous feedback.
See Appendix \ref{app_survey} for details.

\paragraph{Participants}We have a total of 33 respondents to our survey \footnote{While this is not a large sample, it is similar to other recent work which surveys non-cisgender or non-binary people: 19 in \citet{dev2021harms} and 35 in \citet{ungless-etal-2023-stereotypes}}.
Further information on participants can be found in Appendix \ref{app_survey}.

\paragraph{Misgendering experiences}
Most survey respondents faced misgendering on social media platforms, and about a fourth faced misgendering in news articles, biographies, and AI-generated content (Figure \ref{fig:survey}).
There were also some write-in domains where participants faced misgendering, such as journal publications, academic presentations, and website profiles.

\subsection{Desired Interventions for Misgendering}
\label{survey_pref}
We present responses for questions on particular interventions (detect, edit, or hide misgendering content) and open-ended feedback on preferred features from automated intervention systems.

\paragraph{Detect, edit, or hide} The desire for detection of misgendering was high across all domains, with more than three-fourths of the participants wanting misgendering to be automatically detected (Figure \ref{fig:survey}).
As for interventions, participants had varied preferences. 
However, participants had more varied preferences for automatic correction of misgendering.
While about two-thirds of the participants wanted misgendering to be automatically corrected in news articles, AI-generated content, and biographies, only half were interested in the auto-correction of misgendering in social media.
Slightly more participants favored hiding or deleting social media content containing misgendering.
Write-in comments shed light on some nuances to consider for what interventions are appropriate in a given situation:
\begin{itemize}[nosep,leftmargin=5mm]
    \item \textit{Only detect}: Some participants noted that they would only be interested in the automatic detection of misgendering, and would not want the content to be corrected or hidden so they could interpret it themselves.
    \item \textit{Intent based}: Some participants noted that they would want intentional misgendering to make a political point to be hidden but otherwise misgendering content to be corrected.
    \item \textit{Source based}: Some participants expressed that they would only like official content to be auto-corrected, such as journals, articles, biographies, etc.
    Others suggested only AI-generated content should be auto-corrected, and it could violate the American First Amendment right to free speech to edit user-generated content (e.g. social media posts).
\end{itemize}

\begin{figure*}[!ht]
    \centering
    \small
        \includegraphics[width=\textwidth]{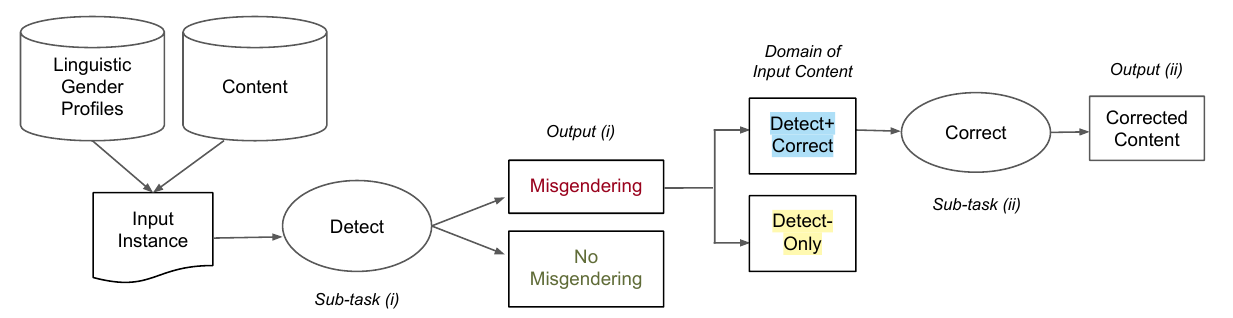}      
    \caption{\textbf{Problem Setup} The misgendering interventions task can be divided into two sub-tasks: (i) detecting misgendering, followed by (ii) correcting misgendering, in domains where editing is appropriate.}
    \label{setup}
\end{figure*}

\noindent Several themes emerge from free-form feedback on desired features for automated interventions: 
\paragraph{Flexible \& user friendly} Any system designed to record individual gender terminology preferences must be customizable (e.g. allow for neo-pronouns) and flexible to modify preferences at any time.
Any misgendering intervention system should operate strictly based on current gender terminology preferences that users have consented to be used for interventions after thorough user education.
It should also be user-friendly, e.g. grammar-correction tools or writing assistants that actively detect and suggest corrections for misgendering during typing.

\paragraph{Conext-sensitivity.} Systems should be sensitive to context in a few different ways:  allowing for different gender terminology in different settings (e.g., neo-pronouns in LGBTQ+ spaces and \textit{they/them} pronouns in non-LGBTQ+ spaces), enabling users to specify different interventions in different domains (e.g. correct misgendering in academic citations but not in job search materials), differentiate between malicious misgendering and unintentional mistakes, and discerning when gender is relevant and when it is not.

\paragraph{LLM fairness \& transparency.}
Language models should have output validation to filter out or correct instances of misgendering. 
They could use proper nouns or default to gender-neutral pronouns such as \textit{they/them} when necessary. 
Reducing the correlation between names, pronouns, professions, personality traits, and physical characteristics in generated content is vital. 
The integration of neo-pronouns and gender-diverse language during the training phase is equally important.
Additionally, there should be transparency about LLM failures and errors regarding misgendering and bias.

\subsection{Concerns about Automated Interventions}
\label{survey_concerns}
There were concerns about the feasibility, limitations, and risks of automated interventions: 

\paragraph{Fundamental infeasibility.} A key concern was that the fluid, flexible, and nuanced nature of individual gender linguistic preferences could not be operationalized.
Any attempt to do so will enforce a static and rigid view of gender in some form.
Simply intervening on text through these systems also would not tackle the root problem of people misgendering others. 

\paragraph{NLP Limitations.} A major concern was that NLP systems are not sophisticated enough to grasp the intricacies of language (e.g. quotations or slang) required for accurate interventions. %
Language models are also biased towards a binary view of gender, stemming from the predominance of binary-gendered language in their training data. 
Addressing this issue is complex; simply removing or altering the binary gendered language in the training corpora is impractical and could compromise their ability to reflect linguistic changes over time.

\paragraph{Censorship and Security.} There is a risk that these systems may unintentionally censor content related to gender-diverse individuals due to errors or overzealous interventions.
There are also several security concerns: these systems could be exploited to target and profile individuals with marginalized and vulnerable gender identities; there could be breaches of privacy, e.g. unintentional \emph{outing} of gender identities; and correcting misgendering might create a mistaken perception of safety and allyship about people who misgender intentionally.

\subsection{Survey Based Dataset Design}

We design our evaluation dataset using insights from the community survey above.
Survey respondents expressed concerns about the potential dangers of automated systems addressing misgendering, such as privacy violations, unintentional disclosure of someone's undisclosed gender identity, or misuse against at-risk groups. 
To minimize risks, we exclusively work with data about public figures who have openly declared their gender identity and gendered terminology preferences.
In any future development of user-oriented intervention systems, such as social media platforms, it is crucial to ensure user autonomy and security.
Key measures include strict adherence to user preferences, secure handling of gender-related information, flexible options for users to opt-in and opt out, and thorough user education about the systems and associated risks, ensuring informed consent at each stage. 

We selected social media and LLM-generations as two domains for our datasets.
We selected social media for several reasons: (i) majority of survey participants experienced misgendering here, (ii)~many respondents showed interest in misgendering detection in this context, (iii) since our focus is on public figures, social media is expected to have relevant posts about them, and (iv) social media platforms offer publicly accessible APIs.
Additionally, we chose LLM-generations as a domain in our dataset because it was a popular domain for both detecting and correcting misgendering, and we can construct instances to challenge the language understanding abilities of NLP systems, thus addressing concerns about their handling of linguistic nuances that were brought up in the survey. %

Further, we implement a source-based separation of interventions, differentiating between \detect and \edit domains.
Social media content is categorized as a \detect domain, aligning with the survey concerns regarding free speech, potential censorship of non-cisgender content, risks of mistaken allyship, and preserving the right to interpret, even potentially offensive, content. 
In contrast, LLM-generated content is designated as a \edit domain, aligned with the interests of survey participants.

\section{\dataset Dataset}
\label{dataset}

\subsection{Problem Setup}
\label{prob_setup}
We assume access to gender profiles on individuals, $P = {p_1, ..., p_{|P|}}$, consisting of their name, gender identity, gender terminology preferences, and deadname, if any. 
The misgendering interventions task can be divided into two sub-tasks: (i) detecting misgendering, followed by (ii) editing misgendering where misgendering is present, in domains where editing is appropriate.
Given a collection of textual content, $C={c_1,..., c_{|C|}}$, about an individual, the first sub-task is to detect, for each input $c$, whether it contains misgendering towards them given their profile $p$.
If so, and if $c$ is from a domain that is appropriate to edit, we continue to the task of editing $c$ to correct the misgendering.
Figure \ref{setup} presents an overview of the problem setup.

\subsection{Data Collection}
\label{data}

We compile a list of notable non-cisgender individuals, including their publicly available gender information. 
We also gather human-written content about them from X and YouTube, as well as text generated by LLMs.

\paragraph{Individuals \& Gender Profiles}
Using the Wikidata Query Service, we extracted the names of individuals identified as 'non-binary', 'trans man', and 'trans woman'. 
We ranked them based on the number of \textit{sitelinks}, which indicate how many Wikipedia pages link to the page about the given individual. 
We focused on the top 10 most popular individuals in each gender category.
For each of these individuals, we used WikiData to gather additional metadata, such as their pronouns and names given at birth.
If an individual's pronouns are missing on WikiData, the pronouns from their Wikipedia biography are used instead.
If a person's name and birth name are different, their birth name is used as their \textit{deadname}\footnote{the name that a transgender person was given at birth and no longer uses upon transitioning \cite{MerriamWebsterDeadname}}.
We inferred appropriate gendered term categories for each individual using their preferred pronouns, utilizing \textit{feminine} terms for those who use \textit{she}, \textit{masculine} terms for \textit{he}, and \textit{neutral} terms for \textit{they}.

\paragraph{X (formerly Twitter) Posts}
We also collected posts from X (formerly Twitter) about each individual using the Twitter API.
If a person's profile consists of a deadname, then we retrieve 50 posts querying for their name and 50 querying for their deadname.
Otherwise, we retrieve 100 posts using their name.
User handles in the text were substituted with \texttt{[USERNAME]} for anonymization, except for those of the relevant public figures.

\paragraph{YouTube Comments} 
We queried the public YouTube Data API using the names and birth names of each individual.
If a person's deadname is available, we queried for 3 videos using their name and 3 videos using their deadname.
Otherwise, we retrieved 6 videos using their name only. 
For each video, we collected 20 comments. 
We also retrieved metadata for both videos and comments.

\paragraph{LLM-Generations} We used GPT-4 \cite{OpenAI2023GPT4TR}, PaLM \cite{Chowdhery2022PaLMSL}, and Vicuna \cite{platzer_et_al:LIPIcs.ECRTS.2021.1} to generate short biographies and sentences about the same group of individuals.
We constructed prompts to generate instances that would challenge the language understanding of NLP systems \cite{ribeiro-etal-2020-beyond} (see Appendix \ref{app_generations} for all prompts).
We split biographies into sentences and annotated per sentence.

\subsection{Annotation}
\label{ann}
Content from all sources is annotated to identify the presence of misgendering. 
We provided Amazon Mechanical Turk (\url{https://www.mturk.com/}) workers with information about each individual (name, gender identities, preferred pronouns, and deadname) along with retrieved texts about them. 
Annotators are asked to label each text instance (YouTube comment, tweet, or generated biography) for whether it contains misgendering towards the query individual (\misg), refers to them without misgendering (\nomisg), or the text is not about the individual (\texttt{Irrelevant}) (Appendix \ref{app_mturk}).
LLM generated text that contains misgendering is also corrected by annotators.

Each instance in our evaluation dataset was annotated by three MTurk workers. 
Workers had to pass a qualification test for each sub-task.
The inter-annotator agreement percentage for detecting misgendering is 87.4\%. 
Conventional agreement scores are unsuitable for correcting misgendering due to the variety of possible valid solutions.
We also did not use human-written edits as gold labels for evaluating baseline models.

We discard instances annotated as \texttt{Irrelevant}.
The \dataset dataset consists of 3790 textual content labeled as \misg or \nomisg towards a paired individual.
LLM-Generations consisting of \misg also consist of human written corrections.
See Table \ref{table_dataset} for a breakdown of the dataset by domain and label.

\paragraph{Challenges}  The first round of annotation instructions, examples, and qualification tests were based on a pilot study (Appendix \ref{pilot}).
However, we noticed annotation errors due to mistaken pronoun coreference resolution (Table \ref{ann_errors}) and updated annotation materials to address this issue.
Annotations using initial guidelines and tests were discarded.

\begin{table}[tb]
\small
    \centering
    \begin{tabularx}{\columnwidth}{X}
    \toprule
\textbf{Tweet:} @USERNAME shes a stalker check out her replies. every ezra miller thread she is there w seething lies who is it? clue [LINK]\\
\textbf{Incorrect Annotation}: \misg\\

\bottomrule
    \end{tabularx}
    \caption{\textbf{Coreference Resolution Error.} Example of an incorrectly annotated tweet about Ezra Miller who uses neutral-gendered words. While the tweet contains feminine pronouns, they are not used to refer to Miller.}
    \label{ann_errors}
\end{table}

\begin{table}[tb]
\centering
\small
\begin{tabular}{@{}lccc@{}}
\toprule
\textbf{Domain} & \textbf{\misg} & \textbf{\nomisg} & \textbf{Total} \\
\midrule
\multirow{2}{*}{X-Posts} & 81 & 1118 & \multirow{2}{*}{1199} \\
 & (6.8\%) & (93.2\%) & \\
\addlinespace
YouTube & 352 & 1217 & \multirow{2}{*}{1559} \\
Comments & (22.0\%) & (78.0\%) & \\
\addlinespace
LLM  & 263 & 769 & \multirow{2}{*}{1032} \\
Generations & (25.5\%) & (74.5\%) & \\
\midrule
\textbf{Grand Total} & & & 3790 \\
\bottomrule
\end{tabular}
\caption{\dataset \textbf{Counts}. Distribution of annotation labels by domain.}
\label{table_dataset}
\end{table}

\subsection{Detect Misgendering}
\label{detect}

\begin{table*}[tb]
\centering
\small
\begin{tabular}{@{}lccccccccc@{}}
\toprule
 & \multicolumn{5}{c}{\textbf{LLM 5-shot CoT}} & \multicolumn{2}{c}{\textbf{Perspective}}& \multicolumn{2}{c}{\textbf{Rule-Based}}\\

\cmidrule(lr){2-6}  %
\cmidrule(lr){7-8}  %
\cmidrule(lr){9-10}  %

          & \textbf{GPT-4} &  \textbf{PaLM} & \textbf{Llama-2} & \textbf{Gemma} & \textbf{Mixtral} & \textbf{Toxicity} & \textbf{Identity} &\textbf{Naive} &\textbf{Coref} \\
\midrule
\textit{X Posts}\\
\hspace{5mm}Accuracy  & \textbf{93.9}  &  86.8   & 59.4 & 70.1  & 56.0 &  91.6     &  79.8      & 77.6              &  87.1   \\
\hspace{5mm}Precision & \textbf{53.5}  &  33.0   & 11.1 & 7.4  & 8 .6 &  12.5     &  15.7      & 22.7               &  26.6   \\
\hspace{5mm}Recall    & 75.3           &  77.8   & 71.6 & 12.3  & 56.8 &  2.5      &  43.2       & \textbf{96.3}    & 51.9    \\
\hspace{5mm}F1        & \textbf{62.6}  &  46.3   & 19.2 & 9.3  & 15.0 &  4.1      &  23.0      & 36.7               &  35.1   \\

\midrule
\textit{YouTube Comments} \\
\hspace{5mm}Accuracy  & \textbf{93.1}  &  85.1 &   64.0 & 60.0& 58.4 & 76.2   &  70.4    &   84.5  & 79.0 \\
\hspace{5mm}Precision & \textbf{80.5}  &  61.0 &  36.7  & 18.8 & 30.4 & 24.0   &  30.6    &   59.2  &  51.2\\
\hspace{5mm}Recall    & 90.6 &            90.6 &  88.6 & 9.1 & 67.5 & 3.5    &  26.6    &   93.9  &  \textbf{94.4}\\
\hspace{5mm}F1        & \textbf{85.3}   & 72.9 &  51.9 & 12.2 & 41.9 & 6.1    &  28.5    &   72.6  &  66.4\\

\midrule
\textit{LLM Generations}\\
\hspace{5mm}Accuracy  &  67.5          &  58.9  &  53.4 & 57.8& 42.0 & 74.5  & 74.5 & 47.7     & \textbf{68.6}  \\
\hspace{5mm}Precision &  42.7          &  36.1  &  31.8 & 22.6& 28.5 & 0.0   & 0.0  &  31.6      &  \textbf{43.1} \\
\hspace{5mm}Recall    &  80.6          &  79.5  &  72.6 & 14.1& 84.0 & 0.0   & 0.0  &  \textbf{90.5}  & 72.2  \\
\hspace{5mm}F1        &  \textbf{55.9} &   49.6 &  44.3 & 17.3& 42.5 &  0.0   & 0.0  & 46.9   &  54.0 \\

\bottomrule
\end{tabular}
\caption{\textbf{Detect results.} Accuracy of the models in detecting \misg in the \dataset dataset.}
\label{detect_results}
\end{table*}

\begin{table*}[tb]
\small
\begin{tabularx}{\textwidth}{llX}
\toprule
\textbf{Error Type}              & \textbf{Profile}                                                                                                            & \textbf{Text, Label, and Prediction}                                                                                                                                                                                                                                                                                                                                                                                                                                                                                                                                                                           \\ \midrule
\textbf{\textit{GPT-4 CoT}}\\
\multirow{3}{*}{Coreference} & \multirow{3}{*}{\begin{tabular}[c]{@{}l@{}}\textbf{Name}: Ezra Miller\\ \textbf{Gendered Terms}: Neutral\end{tabular}} & \textbf{Tweet:} @USERNAME Damn, if Ezra Miller wasn’t a f*****g psychopath then he would make a decent Flash. \\
& & \textbf{Label:} \misg; \textbf{Pred:} \nomisg \\
&  & \textbf{GPT-4 Explanation:} The tweet refers to Ezra Miller but no masculine or feminine words are used to refer to them so there is no misgendering. \\
\cmidrule{2-3}
\multirow{3}{*}{Quotation} & \multirow{3}{*}{\begin{tabular}[c]{@{}l@{}}\textbf{Name:} Elliot Page\\ \textbf{Gendered Terms:} Feminine\\ \textbf{Deadname:}\\\hspace{5mm} Ellen Grace \\\hspace{5mm}Philpotts-Page\end{tabular}} & \textbf{Tweet:} @USERNAME principle of law, or indeed science or nature. "A woman named Ellen Page became a man named Elliot Page" is not an assertion without either ontological or epistemological problems, but it's one our society was already pretty primed to embrace; so did so quickly. \\
                           &                                                                                                                          & \textbf{Label:} \nomisg; \textbf{Pred:} \misg \\
                           &                                                                                                                          & \textbf{GPT-4 Explanation:} The tweet misgenders Elliot Page by calling him a "woman" which is a feminine gendered term, and thus incorrect as Elliot uses neutral and masculine gendered terms. Also, his old name 'Ellen' is used which is his deadname. \\
\cmidrule{2-3}
\multirow{3}{*}{Question}  & \multirow{3}{*}{\begin{tabular}[c]{@{}l@{}}\textbf{Name:} Buck Angel\\ \textbf{Gendered Terms:} Feminine\end{tabular}}                       & \textbf{Tweet:} @USERNAME Do you think Buck Angel should go to the womens changing room? {[}LINK{]} \\
                           &                                                                                                                          & \textbf{Label:} \nomisg; \textbf{Pred:} \misg \\
                           &                                                                                                                          & \textbf{GPT-4 Explanation:} The tweet misgenders Buck Angel by suggesting he should use the women's changing room, implying that he is a woman, this is contrary to Buck's masculine gender identity.  \\
\cmidrule{2-3}
\multirow{3}{*}{Temporal}  & \multirow{3}{*}{\begin{tabular}[c]{@{}l@{}}\textbf{Name:} Lil Uzi Vert\\ \textbf{Gendered Terms:} Neutral\\ \textbf{Deadname:}\\\hspace{5mm} Symere Bysil Wood\end{tabular}}          & \textbf{LLM Generation:} While I used to think of Lil Uzi Vert as a conventional "he-man," I do not think so anymore. \\
                           &                                                                                                                          & \textbf{Label:} \nomisg; \textbf{Pred:} \misg \\
                           &                                                                                                                          & \textbf{GPT-4 Explanation:} The sentence contains the masculine gendered term "he-man" to refer to Lil Uzi Vert... \\
\midrule
\textbf{\textit{Perspective}} \\
\addlinespace
\multirow{3}{*}{\begin{tabular}{@{}l@{}}No toxicity \\ besides \\ misgendering\end{tabular}}
 & \multirow{3}{*}{\begin{tabular}[c]{@{}l@{}}\textbf{Name:} Chaz Bono\\ \textbf{Gendered Terms:} Masculine\\ \textbf{Deadname:} Chastity Bono\end{tabular}} & \textbf{YouTube Comment:} It's a she!! \\
                           &                                                                                                                          & \textbf{Label:} \misg; \textbf{Pred: }\nomisg \\
                           &                                                                                                                          & \\
\midrule

\textbf{\textit{Naive Rule}}\\
\multirow{3}{*}{Typo}     & \multirow{3}{*}{\begin{tabular}[c]{@{}l@{}}\textbf{Name:} Lil Uzi Vert\\ \textbf{Gendered Terms:} Neutral\\ \textbf{Deadname:}\\\hspace{5mm}Symere Bysil Wood\end{tabular}}          & \textbf{LLM Generation:} Lil Uzi Vert is a renowned broither in the rap industry, consistently producing chart-topping hits. \\
                           &                                                                                                                          & \textbf{Label:} \misg; \textbf{Pred:} \nomisg \\
                           &                                                                                                                          & \\
\cmidrule{2-3}
\multirow{3}{*}{Coreference} & \multirow{3}{*}{\begin{tabular}[c]{@{}l@{}}\textbf{Name:} Chaz Bono\\ \textbf{Gendered Terms:} Masculine\\ \textbf{Deadname:} Chastity Bono\end{tabular}} & \textbf{YouTube Comment:} Chaz is a lovely man with a deep understanding of woman's difficulties! \\
                           &                                                                                                                          & \textbf{Label:} \nomisg; \textbf{Pred:} \misg \\
\bottomrule
\end{tabularx}
\caption{\textbf{Detect Errors}. We present examples of instances where benchmark models for detecting misgendering in the \dataset dataset fail at predicting the correct label.}
\label{detect_errors}
\end{table*}

We evaluate several existing NLP tools for detecting misgendering in both \detect and \edit domains.

\paragraph{Prompting} We prompt GPT-4 \cite{OpenAI2023GPT4TR}, PaLM \cite{Chowdhery2022PaLMSL}, Llama-2-Chat 70B \cite{Touvron2023Llama2O}, Gemma-7B-IT \cite{team2024gemma} and Mixtral-8x7B-Instruct \cite{jiang2024mixtral}  with instructions for detecting misgendering with instructions and 5-shot chain-of-thought \cite{Wei2022ChainOT} examples (Appendix \ref{app_detect_prompt}).
For each instance, the person's gender linguistic profile is provided in the prompt as a reference for detecting misgendering, similar to providing evidence sets to verify a claim in fact-checking \cite{gao-etal-2023-rarr}.
Examples are based on instances of misgendering seen in a pilot study (see Appendix \ref{pilot}).

\paragraph{Toxicity Detection} We used the perspective API \cite{Lees2022ANG} for to get scores for toxicity detection and identity attacks.
A threshold of 0.75 was chosen based on a pilot study (Appendix \ref{pilot}) to classify any text with a score above the threshold as containing \misg.

\paragraph{Rule-based} We use a table of pronouns \cite{hossain-etal-2023-misgendered} and a table of gendered keywords created using a list of gendered words from \citet{bolukbasi2016man} (Appendix \ref{app_rule_words}).
For the \textit{naive} approach, if any deadname, gendered word, or pronoun that is inappropriate for a person given their gender linguistic profile (e.g. masculine terms for someone who only uses feminine terminology) is present in the text, then it is classified as containing \misg.
For a \textit{coreference} based approach, \texttt{fastcoref} \cite{Otmazgin2022FcorefFA} is used to create coreference clusters, and if (i) the person's deadname is present in the text, or (ii) an inappropriate gendered word or pronoun is in the same coreference cluster as the person's name or deadname then the instance is predicted to contain \misg.

\paragraph{Results} Across all three data sources we see the highest F1-score for GPT-4 (Table \ref{detect_results}).
While GPT-4 also had the highest precision for X posts and YouTube comments, rule-based methods had the highest recall across all sources.
GPT-4 made errors based on mistaken coreference resolution, and inability to understand some linguistic nuances, such as quotations, questions, and temporal relationships (Table \ref{detect_errors}).
The Perspective API could only positively identify cases of misgendering that were also paired with other forms of toxicity.
Consequently, it could not identify any cases of misgendering in the polite and formal LLM-generated texts.
While the coreference-based method provided the highest precision for LLM-generated misgendering detection, it often failed to create appropriate coreference clusters across data sources.
See Table \ref{detect_errors} for examples of errors from each method.

\subsection{Edit Misgendering}
\label{edit}
We evaluate a few existing NLP tools on their ability to edit misgendering.
Only instances from the \edit domain, LLM-generations, containing \misg are included here. %

\paragraph{Prompting} We prompt GPT-4 , PaLM, and Llama-2-Chat 70B  with instructions for editing misgendering.
For each instance, the individual's gender terminology preferences are provided as a reference, similar to work in non-factual text correction \cite{gao-etal-2023-rarr} (Appendix \ref{app_edit_prompt}).

\paragraph{Rule-based} We create a table gendered words using a list from \citet{bolukbasi2016man} (Appendix \ref{app_rule_words}), and use a table of pronouns from \citet{hossain-etal-2023-misgendered}.
Given a person's gender linguistic profile, if a gendered term or pronoun that is inappropriate for them from these tables is identified in the text, then it is replaced with a corresponding word that matches their linguistic profile.
If switching from a binary pronoun to a neutral one, then corresponding verbs are pluralized \cite{APA2023SingularThey} (Table \ref{edit_alg}).

\paragraph{Results} 
The edited texts were evaluated using human annotators from Amazon Mechanical Turk.
Annotators were asked to evaluate each edited sentence for (i) whether misgendering was corrected, and (ii) whether any unnecessary edits were made.
Three annotators evaluated each instance with an agreement score of 96.3\% for (i) and 89.9\% for (ii).
Due to annotation costs, we only evaluated systems that showed the best performance for detecting misgendering: GPT-4 and the rule-based baseline.
GPT-4 edits corrected misgendering in 97\% of edits, while making unnecessary edits in only 4.6\% of cases.  (Table \ref{edit_eval}).
Unnecessary edits sometimes radically change the original text (Table \ref{edit_errors}).
On the other hand, rule-based baseline corrected misgendering in 78.7\% of the instances, while making unnecessary edits in 28.1\% instances.

\begin{table}[!h]
    \centering
    \small
    \begin{tabular}{ccc}
    \toprule
     \multirow{2}{*}{\textbf{Method} }   & \begin{tabular}[c]{@{}c@{}}\textbf{Misgendering}\\ \textbf{Corrected}\end{tabular} & \begin{tabular}[c]{@{}c@{}}\textbf{Unnecessary}\\ \textbf{Edits}\end{tabular}  \\
     \midrule
      GPT-4   & 97.0 & 4.6 \\
      Rule-based & 78.7 & 28.1 \\
      \bottomrule
    \end{tabular}
    \caption{Human evaluation of automated edits. Edits are evaluated for (i) whether misgendering was corrected, and (ii) whether any unnecessary edits were made.}
    \label{edit_eval}
\end{table}

\begin{table*}[]
\small
\setlength{\tabcolsep}{2pt}
\begin{tabularx}{\textwidth}{lX}
\toprule
\textbf{Profile}   & \textbf{Text, Edits, Labels}  \\ \midrule
\multirow{6}{*}{\begin{tabular}[c]{@{}l@{}}\textbf{Name:} Elliot Page\\ \textbf{Gendered Terms:}\\\hspace{5mm}Feminine\\ \textbf{Deadname:}\\  \hspace{5mm}Ellen Grace \\\hspace{5mm}Philpotts-Page\end{tabular}} &
\textbf{LLM Generation:} She was born in 1846 and spent most of her life as a missionary in China, where she dedicated herself to helping women and children.\\
\addlinespace
&\textbf{GPT-4:} Elliot Page, a renowned Canadian actor and producer, cherishes the special bond they formed with their on-screen mother, Jennifer Garner, while filming the acclaimed movie "Juno". He was born in 1846 and spent most of his life as a missionary in China, where he dedicated himself to helping women and children. \textbf{Label}: \nomisg; \textbf{Unnecessary Edits}: \texttt{Yes}\\
\midrule
\multirow{6}{*}{\begin{tabular}[c]{@{}l@{}}\textbf{Name}: Andreja Pejić\\ \textbf{Gendered Terms:}\\\hspace{5mm}Feminine\\ \textbf{Deadname:}\\\hspace{5mm} Andrej Pejic \end{tabular}} & \textbf{LLM Generation:} While some people might think of Andrej Pejic as a man, I do not agree with this. \\
\addlinespace
&  \textbf{GPT-4:} While some people might think of Andreja Pejić as a man, I do not agree with this. \\
&\textbf{Label:} \nomisg; \textbf{Unnecessary Edits:} No\\
\addlinespace
& \textbf{Rule-based:} While some people might think of Andreja Pejić as a woman, I do not agree with this.\\
& \textbf{Label:} \misg; \textbf{Unnecessary Edits:} No\\

\midrule

\multirow{5}{*}{\begin{tabular}[c]{@{}l@{}}\textbf{Name}: Andreas Krieger\\ \textbf{Gendered Terms:}\\\hspace{5mm} Masculine\\ \textbf{Deadname:}\\\hspace{5mm} Heidi Krieger \end{tabular}} & \textbf{LLM Generation:} This feminine figure is central to Krieger's art, as she seeks to challenge traditional gender norms and celebrate the unique qualities of women. \\
\addlinespace
&  \textbf{GPT-4 Edit}  This masculine figure is central to Krieger's story, as he seeks to challenge traditional gender norms and celebrate his unique journey. \textbf{Label:} \nomisg; \textbf{Unnecessary Edits:} Yes\\
\addlinespace
\bottomrule
\end{tabularx}
\caption{\textbf{Model Edit Examples. } We present examples of instances of LLM generations containing \misg that are edited by GPT-4 or a rule-based editor.
Human annotated labels of the automated edits for whether (i)~whether they still contain misgendering, and (ii) any unnecessary edits were made are also presented.}
\label{edit_errors}
\end{table*}

\section{Related Work}

\paragraph{Gender Bias} 
Significant efforts have been made to address gender bias in language technologies, primarily focusing on a binary and cisgender perspective~\cite{bolukbasi2016man,zhao-etal-2018-gender,kurita2019quantifying}, with recent studies beginning to explore this issue with a non-binary and non-cisgender framework.
\citet{dev2021harms} discuss ways in which gender-exclusivity in NLP can harm non-binary individuals, and demonstrate bias in word embeddings.
\citet{hossain-etal-2023-misgendered} show that LMs are limited in their ability to use non-binary pronouns, \citet{ovalle2023m} evaluate LMs for misgendering and harmful responses to gender disclose, \citet{brandl2022conservative} show neo-pronouns have high perplexity in LMs, \citet{cao-daume-iii-2020-toward} create specialized datasets for coreference resolutions with neo-pronouns, and \citet{lauscher2022welcome} provide desiderata for modeling pronouns in language technologies.
\citet{Sun2021TheyTT} show how models can be trained to rewrite binary pronouns as gender-neutral ones.
While \citet{Lund2023GenderInclusiveGE} introduce a technique to generate singular \textit{they} data and show that data augmentation can mitigate bias against singular \textit{they} in Grammatical Error Correction (GEC) systems.
However, none of these detect and edit misgendering towards given gendered terminology preferences in non-templated texts.

\paragraph{Toxicity Detection and Mitigation} Supervised methods have been extensively used in toxicity detection \cite{Lees2022ANG, kirk-etal-2022-data, fortuna-etal-2022-directions, caselli-etal-2021-hatebert, Poletto2020ResourcesAB}.
Prompted language models have also been used for detecting toxicity in text \cite{Chiu2021DetectingHS, Schick2021SelfDiagnosisAS, goldzycher-schneider-2022-hypothesis}.
\cite{Hallinan2022DetoxifyingTW, ma-etal-2020-powertransformer, malmi-etal-2020-unsupervised} re-write detoxified text using unsupervised masking and reconstruction approaches.
\citet{dale-etal-2021-text, nogueira-dos-santos-etal-2018-fighting} use translation or paraphrasing to detoxify text.
However, none of these works address misgendering as a form of toxicity.

\paragraph{Fact Checking and Correction} Fact-checking is often framed as the task of identifying whether a claim is supported or refuted by the given evidence \cite{wadden-etal-2020-fact, augenstein-etal-2019-multifc, thorne-etal-2018-fever, wang-2017-liar}.
Thre is also work on correcting text that is inconsistent with a set of evidence via post-hoc editing \cite{gao-etal-2023-rarr, iv-etal-2022-fruit, Schick2022PEERAC, thorne-vlachos-2021-evidence}.
However, none of these address misgendering as a form of non-factual information that requires detection and correction.

\section{Conclusion}
In response to the lack of research on automated solutions for misgendering, we conducted a survey among gender-diverse individuals to gather their views on the matter, and based on their responses defined a misgendering interventions task and developed a corresponding evaluation dataset, \dataset.
We provide initial benchmarks for detecting and editing misgendering on this dataset using current NLP systems.
For detecting misgendering, few-shot chain-of-thought prompting of GPT-4 with similar instructions as provided to human annotators achieved the highest F1-score across all data sources (\textit{X posts}: 62.6, \textit{YouTube Comments}: 85.3,\textit{ LLM-generations}: 55.9), but were low enough to indicate significant room for improvement.
Open-source models lagged much further behind with a highest F1-score of 51.9 and a lowest of a mere 9.3.

For the task of correcting misgendering, GPT-4 successfully fixed 97\% of misgendering errors in language model-generated text, with only 4.6\% of edits being unnecessary, as assessed by human annotators. 
However, further work is required as these edits were mainly limited to single, context-free sentences.
For future work, we recommend engaging in wider collaboration with gender-diverse folks to build robust interventions in line with the needs and concerns of the communities most impacted by them.
To facilitate further research, we release the full dataset, code, and demo of our work at \url{https://tamannahossainkay.github.io/misgendermender/}.

\section*{Acknowledgements}
We would like to thank the Queer in AI and the International Society of Non-binary
Scientists (ISBNS) communities, as well as all our anonymous survey participants.
We also thank Yasaman Razeghi, Anthony Chen, Kolby Nottingham, Shivanshu Gupta, Preethi Seshadri, Catarina Belem, Yu Fei, Vinodkumar Prabhakaran, Kathy Meier-Hellstern, Matt Gardner, Yanai Elazar, and anonymous reviewers for their discussions and feedback. 
This work was funded in part by Hasso Plattner Institute (HPI) through the UCI-HPI fellowship, in part by NSF awards IIS-2046873 and IIS-2040989.

\clearpage

\section*{Limitations}

The work in this paper is limited to a Western conception of gender and restricted to English only.

\paragraph{Survey}
This study, though comparable in scale to previous surveys targeting gender-diverse populations, lacks sufficient size for statistically significant findings. 
Our focus was on qualitative evaluation, capturing a range of perspectives within this group. 
However, its limitation to U.S. participants and small sample size impact its generalizability.
To inform the development of effective intervention systems, future research should involve more expansive and comprehensive surveys of gender-diverse individuals.

\paragraph{Task and Dataset}
Our evaluation dataset, featuring publicly available data on public figures, is designed strictly for research purposes. 
It is essential to obtain explicit consent before using this information in any system, and future system development must include informed consent from all human subjects involved.

Our dataset includes prominent public figures who have publicly identified as non-binary, trans men, or trans women, representing only a limited segment of gender identities.
Likewise, the preferred pronouns in the dataset are limited to \textit{she, he} and \textit{they}, with no neo-pronoun representation.
The gender data reflects information available at the time of research and does not account for possible changes thereafter. 

Additionally, the scope of our dataset was confined to social media platforms with accessible APIs and generations from a limited number of LLMs. 
It is important to note that this study does not encompass other text domains where misgendering occurs, such as news articles, biographies, and journals, which remain areas for future research.
LLM generations also contain hallucinations other than misgendering that are not address in this work.
Lastly, to benchmark detection and correction models we use content verified to pertain to a specific individual by human annotators.
In practice, intervention systems would also need to evaluate automated retrieval methods.

\section*{Ethics Statement}
Our research aims to address a particular type of misgendering harm by developing a framework that identifies and amends misgendering in specific settings. 
The work we have published is intended solely for research and should not be employed in the development of any production systems. 
Our community survey is anonymous to safeguard participant identities, and no efforts must be made to identify individual respondents. 
The evaluation dataset we present utilizes publicly accessible information about public figures, exclusively for research objectives. 
It is crucial that this information not be used in any systems without obtaining their explicit consent.

We strictly prohibit using our work for any application that does not have the informed consent of any human subjects involved.
We strictly prohibit the use of our work for censorship, profiling, targeting specific individuals or groups, predicting personal gender identities or terms, or any harmful purposes, particularly against marginalized communities. 
Integral to the future development of such intervention systems is their collaborative creation with the individuals and communities they affect, while ensuring user agency.  
Key measures include secure management of gender-related data, offering users clear options to participate or withdraw, strict compliance with user preferences, and comprehensive user education about the process and its potential risks, ensuring informed consent throughout.

\bibliography{anthology,custom}

\begin{thebibliography}{48}
\expandafter\ifx\csname natexlab\endcsname\relax\def\natexlab#1{#1}\fi

\bibitem[{{APA}(2023)}]{APA2023SingularThey}
{APA}. 2023.
\newblock \href {https://apastyle.apa.org/style-grammar-guidelines/grammar/singular-they} {Grammar and singular they}.
\newblock [Online; accessed 28-November-2023].

\bibitem[{Augenstein et~al.(2019)Augenstein, Lioma, Wang, Chaves~Lima, Hansen, Hansen, and Simonsen}]{augenstein-etal-2019-multifc}
Isabelle Augenstein, Christina Lioma, Dongsheng Wang, Lucas Chaves~Lima, Casper Hansen, Christian Hansen, and Jakob~Grue Simonsen. 2019.
\newblock \href {https://doi.org/10.18653/v1/D19-1475} {{M}ulti{FC}: A real-world multi-domain dataset for evidence-based fact checking of claims}.
\newblock In \emph{Proceedings of the 2019 Conference on Empirical Methods in Natural Language Processing and the 9th International Joint Conference on Natural Language Processing (EMNLP-IJCNLP)}, pages 4685--4697, Hong Kong, China. Association for Computational Linguistics.

\bibitem[{Bolukbasi et~al.(2016)Bolukbasi, Chang, Zou, Saligrama, and Kalai}]{bolukbasi2016man}
Tolga Bolukbasi, Kai{-}Wei Chang, James~Y. Zou, Venkatesh Saligrama, and Adam~Tauman Kalai. 2016.
\newblock \href {https://proceedings.neurips.cc/paper/2016/hash/a486cd07e4ac3d270571622f4f316ec5-Abstract.html} {Man is to computer programmer as woman is to homemaker? debiasing word embeddings}.
\newblock In \emph{Advances in Neural Information Processing Systems 29: Annual Conference on Neural Information Processing Systems 2016, December 5-10, 2016, Barcelona, Spain}, pages 4349--4357.

\bibitem[{Brandl et~al.(2022)Brandl, Cui, and S{\o}gaard}]{brandl2022conservative}
Stephanie Brandl, Ruixiang Cui, and Anders S{\o}gaard. 2022.
\newblock \href {https://doi.org/10.18653/v1/2022.naacl-main.265} {How conservative are language models? adapting to the introduction of gender-neutral pronouns}.
\newblock In \emph{Proceedings of the 2022 Conference of the North American Chapter of the Association for Computational Linguistics: Human Language Technologies}, pages 3624--3630, Seattle, United States. Association for Computational Linguistics.

\bibitem[{Cao and Daum{\'e}~III(2020)}]{cao-daume-iii-2020-toward}
Yang~Trista Cao and Hal Daum{\'e}~III. 2020.
\newblock \href {https://doi.org/10.18653/v1/2020.acl-main.418} {Toward gender-inclusive coreference resolution}.
\newblock In \emph{Proceedings of the 58th Annual Meeting of the Association for Computational Linguistics}, pages 4568--4595, Online. Association for Computational Linguistics.

\bibitem[{Caselli et~al.(2021)Caselli, Basile, Mitrovi{\'c}, and Granitzer}]{caselli-etal-2021-hatebert}
Tommaso Caselli, Valerio Basile, Jelena Mitrovi{\'c}, and Michael Granitzer. 2021.
\newblock \href {https://doi.org/10.18653/v1/2021.woah-1.3} {{H}ate{BERT}: Retraining {BERT} for abusive language detection in {E}nglish}.
\newblock In \emph{Proceedings of the 5th Workshop on Online Abuse and Harms (WOAH 2021)}, pages 17--25, Online. Association for Computational Linguistics.

\bibitem[{Chiu and Alexander(2021)}]{Chiu2021DetectingHS}
Ke-Li Chiu and Rohan Alexander. 2021.
\newblock \href {https://arxiv.org/abs/2103.12407} {Detecting hate speech with gpt-3}.
\newblock \emph{ArXiv preprint}, abs/2103.12407.

\bibitem[{Choubey et~al.(2021)Choubey, Currey, Mathur, and Dinu}]{choubey-etal-2021-gfst}
Prafulla~Kumar Choubey, Anna Currey, Prashant Mathur, and Georgiana Dinu. 2021.
\newblock \href {https://doi.org/10.18653/v1/2021.emnlp-main.123} {{GFST}: {G}ender-filtered self-training for more accurate gender in translation}.
\newblock In \emph{Proceedings of the 2021 Conference on Empirical Methods in Natural Language Processing}, pages 1640--1654, Online and Punta Cana, Dominican Republic. Association for Computational Linguistics.

\bibitem[{Chowdhery et~al.(2022)Chowdhery, Narang, Devlin, Bosma, Mishra, Roberts, Barham, Chung, Sutton, Gehrmann, Schuh, Shi, Tsvyashchenko, Maynez, Rao, Barnes, Tay, Shazeer, Prabhakaran, Reif, Du, Hutchinson, Pope, Bradbury, Austin, Isard, Gur-Ari, Yin, Duke, Levskaya, Ghemawat, Dev, Michalewski, Garc{\'i}a, Misra, Robinson, Fedus, Zhou, Ippolito, Luan, Lim, Zoph, Spiridonov, Sepassi, Dohan, Agrawal, Omernick, Dai, Pillai, Pellat, Lewkowycz, Moreira, Child, Polozov, Lee, Zhou, Wang, Saeta, D{\'i}az, Firat, Catasta, Wei, Meier-Hellstern, Eck, Dean, Petrov, and Fiedel}]{Chowdhery2022PaLMSL}
Aakanksha Chowdhery, Sharan Narang, Jacob Devlin, Maarten Bosma, Gaurav Mishra, Adam Roberts, Paul Barham, Hyung~Won Chung, Charles Sutton, Sebastian Gehrmann, Parker Schuh, Kensen Shi, Sasha Tsvyashchenko, Joshua Maynez, Abhishek Rao, Parker Barnes, Yi~Tay, Noam~M. Shazeer, Vinodkumar Prabhakaran, Emily Reif, Nan Du, Benton~C. Hutchinson, Reiner Pope, James Bradbury, Jacob Austin, Michael Isard, Guy Gur-Ari, Pengcheng Yin, Toju Duke, Anselm Levskaya, Sanjay Ghemawat, Sunipa Dev, Henryk Michalewski, Xavier Garc{\'i}a, Vedant Misra, Kevin Robinson, Liam Fedus, Denny Zhou, Daphne Ippolito, David Luan, Hyeontaek Lim, Barret Zoph, Alexander Spiridonov, Ryan Sepassi, David Dohan, Shivani Agrawal, Mark Omernick, Andrew~M. Dai, Thanumalayan~Sankaranarayana Pillai, Marie Pellat, Aitor Lewkowycz, Erica Moreira, Rewon Child, Oleksandr Polozov, Katherine Lee, Zongwei Zhou, Xuezhi Wang, Brennan Saeta, Mark D{\'i}az, Orhan Firat, Michele Catasta, Jason Wei, Kathleen~S. Meier-Hellstern, Douglas Eck, Jeff Dean, Slav Petrov,
  and Noah Fiedel. 2022.
\newblock \href {https://api.semanticscholar.org/CorpusID:247951931} {Palm: Scaling language modeling with pathways}.
\newblock \emph{J. Mach. Learn. Res.}, 24:240:1--240:113.

\bibitem[{Dale et~al.(2021)Dale, Voronov, Dementieva, Logacheva, Kozlova, Semenov, and Panchenko}]{dale-etal-2021-text}
David Dale, Anton Voronov, Daryna Dementieva, Varvara Logacheva, Olga Kozlova, Nikita Semenov, and Alexander Panchenko. 2021.
\newblock \href {https://doi.org/10.18653/v1/2021.emnlp-main.629} {Text detoxification using large pre-trained neural models}.
\newblock In \emph{Proceedings of the 2021 Conference on Empirical Methods in Natural Language Processing}, pages 7979--7996, Online and Punta Cana, Dominican Republic. Association for Computational Linguistics.

\bibitem[{Dev et~al.(2021)Dev, Monajatipoor, Ovalle, Subramonian, Phillips, and Chang}]{dev2021harms}
Sunipa Dev, Masoud Monajatipoor, Anaelia Ovalle, Arjun Subramonian, Jeff Phillips, and Kai-Wei Chang. 2021.
\newblock \href {https://doi.org/10.18653/v1/2021.emnlp-main.150} {Harms of gender exclusivity and challenges in non-binary representation in language technologies}.
\newblock In \emph{Proceedings of the 2021 Conference on Empirical Methods in Natural Language Processing}, pages 1968--1994, Online and Punta Cana, Dominican Republic. Association for Computational Linguistics.

\bibitem[{Dictionary(2023)}]{oed_misgender}
Oxford~English Dictionary. 2023.
\newblock \href {https://www.oed.com/dictionary/misgender_v?tab=meaning_and_use#1223410870} {Definition of misgender}.
\newblock Accessed: 2023-10-06.

\bibitem[{Fortuna et~al.(2022)Fortuna, Dominguez, Wanner, and Talat}]{fortuna-etal-2022-directions}
Paula Fortuna, Monica Dominguez, Leo Wanner, and Zeerak Talat. 2022.
\newblock \href {https://doi.org/10.18653/v1/2022.emnlp-main.809} {Directions for {NLP} practices applied to online hate speech detection}.
\newblock In \emph{Proceedings of the 2022 Conference on Empirical Methods in Natural Language Processing}, pages 11794--11805, Abu Dhabi, United Arab Emirates. Association for Computational Linguistics.

\bibitem[{Gao et~al.(2023)Gao, Dai, Pasupat, Chen, Chaganty, Fan, Zhao, Lao, Lee, Juan, and Guu}]{gao-etal-2023-rarr}
Luyu Gao, Zhuyun Dai, Panupong Pasupat, Anthony Chen, Arun~Tejasvi Chaganty, Yicheng Fan, Vincent Zhao, Ni~Lao, Hongrae Lee, Da-Cheng Juan, and Kelvin Guu. 2023.
\newblock \href {https://doi.org/10.18653/v1/2023.acl-long.910} {{RARR}: Researching and revising what language models say, using language models}.
\newblock In \emph{Proceedings of the 61st Annual Meeting of the Association for Computational Linguistics (Volume 1: Long Papers)}, pages 16477--16508, Toronto, Canada. Association for Computational Linguistics.

\bibitem[{Gao and Emami(2023)}]{gao-emami-2023-turing}
Qi~Chen Gao and Ali Emami. 2023.
\newblock \href {https://doi.org/10.18653/v1/2023.acl-srw.17} {The {T}uring quest: Can transformers make good {NPC}s?}
\newblock In \emph{Proceedings of the 61st Annual Meeting of the Association for Computational Linguistics (Volume 4: Student Research Workshop)}, pages 93--103, Toronto, Canada. Association for Computational Linguistics.

\bibitem[{Goldzycher and Schneider(2022)}]{goldzycher-schneider-2022-hypothesis}
Janis Goldzycher and Gerold Schneider. 2022.
\newblock \href {https://aclanthology.org/2022.trac-1.10} {Hypothesis engineering for zero-shot hate speech detection}.
\newblock In \emph{Proceedings of the Third Workshop on Threat, Aggression and Cyberbullying (TRAC 2022)}, pages 75--90, Gyeongju, Republic of Korea. Association for Computational Linguistics.

\bibitem[{Guo et~al.(2022)Guo, Yang, and Abbasi}]{guo-etal-2022-auto}
Yue Guo, Yi~Yang, and Ahmed Abbasi. 2022.
\newblock \href {https://doi.org/10.18653/v1/2022.acl-long.72} {Auto-debias: Debiasing masked language models with automated biased prompts}.
\newblock In \emph{Proceedings of the 60th Annual Meeting of the Association for Computational Linguistics (Volume 1: Long Papers)}, pages 1012--1023, Dublin, Ireland. Association for Computational Linguistics.

\bibitem[{Hallinan et~al.(2022)Hallinan, Liu, Choi, and Sap}]{Hallinan2022DetoxifyingTW}
Skyler Hallinan, Alisa Liu, Yejin Choi, and Maarten Sap. 2022.
\newblock \href {https://api.semanticscholar.org/CorpusID:252734135} {Detoxifying text with marco: Controllable revision with experts and anti-experts}.
\newblock In \emph{Annual Meeting of the Association for Computational Linguistics}.

\bibitem[{Hossain et~al.(2023)Hossain, Dev, and Singh}]{hossain-etal-2023-misgendered}
Tamanna Hossain, Sunipa Dev, and Sameer Singh. 2023.
\newblock \href {https://doi.org/10.18653/v1/2023.acl-long.293} {{MISGENDERED}: Limits of large language models in understanding pronouns}.
\newblock In \emph{Proceedings of the 61st Annual Meeting of the Association for Computational Linguistics (Volume 1: Long Papers)}, pages 5352--5367, Toronto, Canada. Association for Computational Linguistics.

\bibitem[{Iv et~al.(2022)Iv, Passos, Singh, and Chang}]{iv-etal-2022-fruit}
Robert Iv, Alexandre Passos, Sameer Singh, and Ming-Wei Chang. 2022.
\newblock \href {https://doi.org/10.18653/v1/2022.naacl-main.269} {{FRUIT}: Faithfully reflecting updated information in text}.
\newblock In \emph{Proceedings of the 2022 Conference of the North American Chapter of the Association for Computational Linguistics: Human Language Technologies}, pages 3670--3686, Seattle, United States. Association for Computational Linguistics.

\bibitem[{Jiang et~al.(2024)Jiang, Sablayrolles, Roux, Mensch, Savary, Bamford, Chaplot, Casas, Hanna, Bressand et~al.}]{jiang2024mixtral}
Albert~Q Jiang, Alexandre Sablayrolles, Antoine Roux, Arthur Mensch, Blanche Savary, Chris Bamford, Devendra~Singh Chaplot, Diego de~las Casas, Emma~Bou Hanna, Florian Bressand, et~al. 2024.
\newblock \href {https://arxiv.org/abs/2401.04088} {Mixtral of experts}.
\newblock \emph{ArXiv preprint}, abs/2401.04088.

\bibitem[{Kirk et~al.(2022)Kirk, Vidgen, and Hale}]{kirk-etal-2022-data}
Hannah Kirk, Bertie Vidgen, and Scott Hale. 2022.
\newblock \href {https://aclanthology.org/2022.trac-1.7} {Is more data better? re-thinking the importance of efficiency in abusive language detection with transformers-based active learning}.
\newblock In \emph{Proceedings of the Third Workshop on Threat, Aggression and Cyberbullying (TRAC 2022)}, pages 52--61, Gyeongju, Republic of Korea. Association for Computational Linguistics.

\bibitem[{Kurita et~al.(2019)Kurita, Vyas, Pareek, Black, and Tsvetkov}]{kurita2019quantifying}
Keita Kurita, Nidhi Vyas, Ayush Pareek, Alan~W Black, and Yulia Tsvetkov. 2019.
\newblock \href {https://doi.org/10.18653/v1/W19-3823} {Measuring bias in contextualized word representations}.
\newblock In \emph{Proceedings of the First Workshop on Gender Bias in Natural Language Processing}, pages 166--172, Florence, Italy. Association for Computational Linguistics.

\bibitem[{Lauscher et~al.(2022)Lauscher, Crowley, and Hovy}]{lauscher2022welcome}
Anne Lauscher, Archie Crowley, and Dirk Hovy. 2022.
\newblock \href {https://aclanthology.org/2022.coling-1.105} {Welcome to the modern world of pronouns: Identity-inclusive natural language processing beyond gender}.
\newblock In \emph{Proceedings of the 29th International Conference on Computational Linguistics}, pages 1221--1232, Gyeongju, Republic of Korea. International Committee on Computational Linguistics.

\bibitem[{Lees et~al.(2022)Lees, Tran, Tay, Sorensen, Gupta, Metzler, and Vasserman}]{Lees2022ANG}
Alyssa Lees, Vinh~Q. Tran, Yi~Tay, Jeffrey~Scott Sorensen, Jai Gupta, Donald Metzler, and Lucy Vasserman. 2022.
\newblock \href {https://api.semanticscholar.org/CorpusID:247058801} {A new generation of perspective api: Efficient multilingual character-level transformers}.
\newblock \emph{Proceedings of the 28th ACM SIGKDD Conference on Knowledge Discovery and Data Mining}.

\bibitem[{Lund et~al.(2023)Lund, Omelianchuk, and Samokhin}]{Lund2023GenderInclusiveGE}
Gunnar Lund, Kostiantyn Omelianchuk, and Igor Samokhin. 2023.
\newblock \href {https://api.semanticscholar.org/CorpusID:259145073} {Gender-inclusive grammatical error correction through augmentation}.
\newblock In \emph{Workshop on Innovative Use of NLP for Building Educational Applications}.

\bibitem[{Ma et~al.(2020)Ma, Sap, Rashkin, and Choi}]{ma-etal-2020-powertransformer}
Xinyao Ma, Maarten Sap, Hannah Rashkin, and Yejin Choi. 2020.
\newblock \href {https://doi.org/10.18653/v1/2020.emnlp-main.602} {{P}ower{T}ransformer: Unsupervised controllable revision for biased language correction}.
\newblock In \emph{Proceedings of the 2020 Conference on Empirical Methods in Natural Language Processing (EMNLP)}, pages 7426--7441, Online. Association for Computational Linguistics.

\bibitem[{Malmi et~al.(2020)Malmi, Severyn, and Rothe}]{malmi-etal-2020-unsupervised}
Eric Malmi, Aliaksei Severyn, and Sascha Rothe. 2020.
\newblock \href {https://doi.org/10.18653/v1/2020.emnlp-main.699} {Unsupervised text style transfer with padded masked language models}.
\newblock In \emph{Proceedings of the 2020 Conference on Empirical Methods in Natural Language Processing (EMNLP)}, pages 8671--8680, Online. Association for Computational Linguistics.

\bibitem[{{Merriam-Webster}(2023)}]{MerriamWebsterDeadname}
{Merriam-Webster}. 2023.
\newblock Definition of deadname.
\newblock \url{https://www.merriam-webster.com/dictionary/deadname}.
\newblock Accessed: 2023-12-14.

\bibitem[{Nogueira~dos Santos et~al.(2018)Nogueira~dos Santos, Melnyk, and Padhi}]{nogueira-dos-santos-etal-2018-fighting}
Cicero Nogueira~dos Santos, Igor Melnyk, and Inkit Padhi. 2018.
\newblock \href {https://doi.org/10.18653/v1/P18-2031} {Fighting offensive language on social media with unsupervised text style transfer}.
\newblock In \emph{Proceedings of the 56th Annual Meeting of the Association for Computational Linguistics (Volume 2: Short Papers)}, pages 189--194, Melbourne, Australia. Association for Computational Linguistics.

\bibitem[{OpenAI(2023)}]{OpenAI2023GPT4TR}
OpenAI. 2023.
\newblock \href {https://arxiv.org/abs/2303.08774} {Gpt-4 technical report}.
\newblock \emph{ArXiv preprint}, abs/2303.08774.

\bibitem[{Otmazgin et~al.(2022)Otmazgin, Cattan, and Goldberg}]{Otmazgin2022FcorefFA}
Shon Otmazgin, Arie Cattan, and Yoav Goldberg. 2022.
\newblock \href {https://aclanthology.org/2022.aacl-demo.6} {{F}-coref: Fast, accurate and easy to use coreference resolution}.
\newblock In \emph{Proceedings of the 2nd Conference of the Asia-Pacific Chapter of the Association for Computational Linguistics and the 12th International Joint Conference on Natural Language Processing: System Demonstrations}, pages 48--56, Taipei, Taiwan. Association for Computational Linguistics.

\bibitem[{Ovalle et~al.(2023)Ovalle, Goyal, Dhamala, Jaggers, Chang, Galstyan, Zemel, and Gupta}]{ovalle2023m}
Anaelia Ovalle, Palash Goyal, Jwala Dhamala, Zachary Jaggers, Kai-Wei Chang, Aram Galstyan, Richard Zemel, and Rahul Gupta. 2023.
\newblock \href {https://arxiv.org/abs/2305.09941} {" i'm fully who i am": Towards centering transgender and non-binary voices to measure biases in open language generation}.
\newblock \emph{ArXiv preprint}, abs/2305.09941.

\bibitem[{Platzer and Puschner(2021)}]{platzer_et_al:LIPIcs.ECRTS.2021.1}
Michael Platzer and Peter Puschner. 2021.
\newblock \href {https://doi.org/10.4230/LIPIcs.ECRTS.2021.1} {{Vicuna: A Timing-Predictable RISC-V Vector Coprocessor for Scalable Parallel Computation}}.
\newblock In \emph{33rd Euromicro Conference on Real-Time Systems (ECRTS 2021)}, volume 196 of \emph{Leibniz International Proceedings in Informatics (LIPIcs)}, pages 1:1--1:18, Dagstuhl, Germany. Schloss Dagstuhl -- Leibniz-Zentrum f{\"u}r Informatik.

\bibitem[{Poletto et~al.(2020)Poletto, Basile, Sanguinetti, Bosco, and Patti}]{Poletto2020ResourcesAB}
Fabio Poletto, Valerio Basile, Manuela Sanguinetti, Cristina Bosco, and Viviana Patti. 2020.
\newblock \href {https://api.semanticscholar.org/CorpusID:224846337} {Resources and benchmark corpora for hate speech detection: a systematic review}.
\newblock \emph{Language Resources and Evaluation}, 55:477 -- 523.

\bibitem[{Ribeiro et~al.(2020)Ribeiro, Wu, Guestrin, and Singh}]{ribeiro-etal-2020-beyond}
Marco~Tulio Ribeiro, Tongshuang Wu, Carlos Guestrin, and Sameer Singh. 2020.
\newblock \href {https://doi.org/10.18653/v1/2020.acl-main.442} {Beyond accuracy: Behavioral testing of {NLP} models with {C}heck{L}ist}.
\newblock In \emph{Proceedings of the 58th Annual Meeting of the Association for Computational Linguistics}, pages 4902--4912, Online. Association for Computational Linguistics.

\bibitem[{Schick et~al.(2022)Schick, Dwivedi-Yu, Jiang, Petroni, Lewis, Izacard, You, Nalmpantis, Grave, and Riedel}]{Schick2022PEERAC}
Timo Schick, Jane Dwivedi-Yu, Zhengbao Jiang, Fabio Petroni, Patrick Lewis, Gautier Izacard, Qingfei You, Christoforos Nalmpantis, Edouard Grave, and Sebastian Riedel. 2022.
\newblock \href {https://arxiv.org/abs/2208.11663} {Peer: A collaborative language model}.
\newblock \emph{ArXiv preprint}, abs/2208.11663.

\bibitem[{Schick et~al.(2021)Schick, Udupa, and Sch{\"u}tze}]{Schick2021SelfDiagnosisAS}
Timo Schick, Sahana Udupa, and Hinrich Sch{\"u}tze. 2021.
\newblock \href {https://doi.org/10.1162/tacl_a_00434} {Self-diagnosis and self-debiasing: A proposal for reducing corpus-based bias in {NLP}}.
\newblock \emph{Transactions of the Association for Computational Linguistics}, 9:1408--1424.

\bibitem[{Sun et~al.(2021)Sun, Webster, Shah, Wang, and Johnson}]{Sun2021TheyTT}
Tony Sun, Kellie Webster, Apurva Shah, William~Yang Wang, and Melvin Johnson. 2021.
\newblock \href {https://arxiv.org/abs/2102.06788} {They, them, theirs: Rewriting with gender-neutral english}.
\newblock \emph{ArXiv preprint}, abs/2102.06788.

\bibitem[{Team et~al.(2024)Team, Mesnard, Hardin, Dadashi, Bhupatiraju, Pathak, Sifre, Rivi{\`e}re, Kale, Love et~al.}]{team2024gemma}
Gemma Team, Thomas Mesnard, Cassidy Hardin, Robert Dadashi, Surya Bhupatiraju, Shreya Pathak, Laurent Sifre, Morgane Rivi{\`e}re, Mihir~Sanjay Kale, Juliette Love, et~al. 2024.
\newblock \href {https://arxiv.org/abs/2403.08295} {Gemma: Open models based on gemini research and technology}.
\newblock \emph{ArXiv preprint}, abs/2403.08295.

\bibitem[{Thorne and Vlachos(2021)}]{thorne-vlachos-2021-evidence}
James Thorne and Andreas Vlachos. 2021.
\newblock \href {https://doi.org/10.18653/v1/2021.acl-long.256} {Evidence-based factual error correction}.
\newblock In \emph{Proceedings of the 59th Annual Meeting of the Association for Computational Linguistics and the 11th International Joint Conference on Natural Language Processing (Volume 1: Long Papers)}, pages 3298--3309, Online. Association for Computational Linguistics.

\bibitem[{Thorne et~al.(2018)Thorne, Vlachos, Christodoulopoulos, and Mittal}]{thorne-etal-2018-fever}
James Thorne, Andreas Vlachos, Christos Christodoulopoulos, and Arpit Mittal. 2018.
\newblock \href {https://doi.org/10.18653/v1/N18-1074} {{FEVER}: a large-scale dataset for fact extraction and {VER}ification}.
\newblock In \emph{Proceedings of the 2018 Conference of the North {A}merican Chapter of the Association for Computational Linguistics: Human Language Technologies, Volume 1 (Long Papers)}, pages 809--819, New Orleans, Louisiana. Association for Computational Linguistics.

\bibitem[{Touvron et~al.(2023)Touvron, Martin, Stone, Albert, Almahairi, Babaei, Bashlykov, Batra, Bhargava, Bhosale, Bikel, Blecher, Ferrer, Chen, Cucurull, Esiobu, Fernandes, Fu, Fu, Fuller, Gao, Goswami, Goyal, Hartshorn, Hosseini, Hou, Inan, Kardas, Kerkez, Khabsa, Kloumann, Korenev, Koura, Lachaux, Lavril, Lee, Liskovich, Lu, Mao, Martinet, Mihaylov, Mishra, Molybog, Nie, Poulton, Reizenstein, Rungta, Saladi, Schelten, Silva, Smith, Subramanian, Tan, Tang, Taylor, Williams, Kuan, Xu, Yan, Zarov, Zhang, Fan, Kambadur, Narang, Rodriguez, Stojnic, Edunov, and Scialom}]{Touvron2023Llama2O}
Hugo Touvron, Louis Martin, Kevin~R. Stone, Peter Albert, Amjad Almahairi, Yasmine Babaei, Nikolay Bashlykov, Soumya Batra, Prajjwal Bhargava, Shruti Bhosale, Daniel~M. Bikel, Lukas Blecher, Cristian~Cant{\'o}n Ferrer, Moya Chen, Guillem Cucurull, David Esiobu, Jude Fernandes, Jeremy Fu, Wenyin Fu, Brian Fuller, Cynthia Gao, Vedanuj Goswami, Naman Goyal, Anthony~S. Hartshorn, Saghar Hosseini, Rui Hou, Hakan Inan, Marcin Kardas, Viktor Kerkez, Madian Khabsa, Isabel~M. Kloumann, A.~V. Korenev, Punit~Singh Koura, Marie-Anne Lachaux, Thibaut Lavril, Jenya Lee, Diana Liskovich, Yinghai Lu, Yuning Mao, Xavier Martinet, Todor Mihaylov, Pushkar Mishra, Igor Molybog, Yixin Nie, Andrew Poulton, Jeremy Reizenstein, Rashi Rungta, Kalyan Saladi, Alan Schelten, Ruan Silva, Eric~Michael Smith, R.~Subramanian, Xia Tan, Binh Tang, Ross Taylor, Adina Williams, Jian~Xiang Kuan, Puxin Xu, Zhengxu Yan, Iliyan Zarov, Yuchen Zhang, Angela Fan, Melanie Kambadur, Sharan Narang, Aurelien Rodriguez, Robert Stojnic, Sergey Edunov, and
  Thomas Scialom. 2023.
\newblock \href {https://arxiv.org/abs/2307.09288} {Llama 2: Open foundation and fine-tuned chat models}.
\newblock \emph{ArXiv preprint}, abs/2307.09288.

\bibitem[{Ungless et~al.(2023)Ungless, Ross, and Lauscher}]{ungless-etal-2023-stereotypes}
Eddie Ungless, Bjorn Ross, and Anne Lauscher. 2023.
\newblock \href {https://doi.org/10.18653/v1/2023.findings-acl.502} {Stereotypes and smut: The (mis)representation of non-cisgender identities by text-to-image models}.
\newblock In \emph{Findings of the Association for Computational Linguistics: ACL 2023}, pages 7919--7942, Toronto, Canada. Association for Computational Linguistics.

\bibitem[{Wadden et~al.(2020)Wadden, Lin, Lo, Wang, van Zuylen, Cohan, and Hajishirzi}]{wadden-etal-2020-fact}
David Wadden, Shanchuan Lin, Kyle Lo, Lucy~Lu Wang, Madeleine van Zuylen, Arman Cohan, and Hannaneh Hajishirzi. 2020.
\newblock \href {https://doi.org/10.18653/v1/2020.emnlp-main.609} {Fact or fiction: Verifying scientific claims}.
\newblock In \emph{Proceedings of the 2020 Conference on Empirical Methods in Natural Language Processing (EMNLP)}, pages 7534--7550, Online. Association for Computational Linguistics.

\bibitem[{Wang(2017)}]{wang-2017-liar}
William~Yang Wang. 2017.
\newblock \href {https://doi.org/10.18653/v1/P17-2067} {{``}liar, liar pants on fire{''}: A new benchmark dataset for fake news detection}.
\newblock In \emph{Proceedings of the 55th Annual Meeting of the Association for Computational Linguistics (Volume 2: Short Papers)}, pages 422--426, Vancouver, Canada. Association for Computational Linguistics.

\bibitem[{Wei et~al.(2022)Wei, Wang, Schuurmans, Bosma, hsin Chi, Xia, Le, and Zhou}]{Wei2022ChainOT}
Jason Wei, Xuezhi Wang, Dale Schuurmans, Maarten Bosma, Ed~Huai hsin Chi, F.~Xia, Quoc Le, and Denny Zhou. 2022.
\newblock \href {https://arxiv.org/abs/2201.11903} {Chain of thought prompting elicits reasoning in large language models}.
\newblock \emph{ArXiv preprint}, abs/2201.11903.

\bibitem[{Zhao et~al.(2018)Zhao, Wang, Yatskar, Ordonez, and Chang}]{zhao-etal-2018-gender}
Jieyu Zhao, Tianlu Wang, Mark Yatskar, Vicente Ordonez, and Kai-Wei Chang. 2018.
\newblock \href {https://doi.org/10.18653/v1/N18-2003} {Gender bias in coreference resolution: Evaluation and debiasing methods}.
\newblock In \emph{Proceedings of the 2018 Conference of the North {A}merican Chapter of the Association for Computational Linguistics: Human Language Technologies, Volume 2 (Short Papers)}, pages 15--20, New Orleans, Louisiana. Association for Computational Linguistics.

\end{thebibliography}

\appendix

\section{Survey}
\label{app_survey}
\subsection{IRB Self-Exempt}

Using the IRB Exempt Self-Determination Tool, our survey was determined to be exempt from IRB review under Category 2 (i) and (ii) \footnote{ https://www.hhs.gov/ohrp/regulations-and-policy/regulations/45-cfr-46/common-rule-subpart-a-46104/index.html}.

\subsection{Informed Consent}

Lead Researcher: \texttt{[NAME]}, Faculty: \texttt{[NAME]}

Please read the information below and ask questions about anything that you do not understand. The lead researcher listed above will be available to answer your questions.

\begin{itemize}
    
\item You are invited to participate in a research study. Participation in this study is voluntary. You may refuse to participate or discontinue your involvement at any time without penalty or loss of benefits. You are free to withdraw from this study at any time.

\item To participate in this study you must be 18 or older, and located in the United States of America.

\item We would find it helpful for you to complete a survey to learn more about how language technologies can identify and address misgendering issues in textual content relating to non-binary and transgender individuals. 

\item The survey consists of 4 short sections and might take ~10 to 15 minutes to complete.

\item No personally identifiable information about participants will be collected as part of this study. Your responses are completely anonymous.

\item Possible risks/discomforts associated with the study are emotional distress from questions about gender misidentification, or the potential triggering of past traumas related to misgendering.

\item There are no direct benefits from participation in the study.  However, this study may contribute to the development of tools aimed at detecting and counteracting misgendering in textual content.

\item Data storage: The information you provide will be collected and stored using Google Forms, a third-party online platform. The data collected via Google Forms will be stored on secure servers managed by Google, in accordance with their data privacy policies.

\item Data Access and Future Use: The lead researcher and team will view the anonymous responses from this study. After the study's conclusion, these responses may be shared with other researchers for future studies. Further permissions for data sharing will not be sought.

\item Questions? If you have any comments, concerns, or questions regarding this study please contact the lead researcher listed at the top of this form.

\item If you have questions or concerns about your rights as a research participant, you can contact the \texttt{[INSTITUTE]} Institutional Review Board by phone, \texttt{[PHONE NUMBER]}, by e-mail at \texttt{[EMAIL]} or at \texttt{[ADDRESS]}.

 What is an IRB?  An Institutional Review Board (IRB) is a committee made up of scientists and non-scientists.  The IRB’s role is to protect the rights and welfare of human subjects involved in research.  The IRB also assures that the research complies with applicable regulations, laws, and institutional policies.

• If you consent to participate in this study, check the box below and start the survey by clicking 'Next'

\end{itemize}

\subsection{Survey Questions}
Below is a description of the survey's four sections, accompanied by their respective questions. The format of each answer - checkboxes\footnote{All checkbox questions have an 'Other' option with a free-form text field to write-in answers.}, radio buttons, or free-form text - is indicated in parentheses next to the questions.

\paragraph{Demographic information} To understand the gender and linguistic diversity of our participants, in this section we ask participants to specify their gender identity and their chosen personal pronouns. 
Additionally, to ensure adherence to the study's criteria, we verify if the participant is an adult and currently residing within the United States.
The questions were as follows:
\begin{itemize}
\item What is your gender identity? (checkboxes)
\item What pronouns do you use? (checkboxes)
\item What is your age group? (radio buttons)
\item What is your country of residence? (radio buttons)
\end{itemize}

\paragraph{Misgendering experiences and desired interventions} To determine where misgendering is prevalent and identify effective interventions, we ask participants whether they have faced misgendering in each of four domains: social media (e.g., Twitter, YouTube), biographies, news articles, and user-generated content, with an option for participants to specify additional domains. 
For each domain, we ask participants to specify whether they would be interested in the following interventions for instances of misgendering: flagging or detecting, automatic corrections, and hiding or removal. 
Additionally, we ask them to describe in which instances would they favor correction instead of hiding or removal and vice versa. 
The questions were as follows:
\begin{itemize}
\item Have you faced misgendering in any of these domains? (checkboxes)
\item Would you want misgendering detected and flagged for users in any of these domains? (checkboxes)
\item Would you want misgendering to be automatically corrected in any of these domains? (checkboxes)
\item Would you want misgendering to be automatically hidden or deleted in any of these domains? (checkboxes)
\item What types of misgendering content would you want automatically corrected vs. hidden/deleted? (free-form text)
\end{itemize}

\paragraph{NLP technologies} To gather insights from across different levels of expertise regarding NLP, we ask participants to rate their familiarity with language technologies from 1(low) to 5 (high), and free-form questions on what functionality would they like to see in language technologies to effectively address misgendering,  as well as potential concerns regarding such technologies. 
The questions were as follows:
\begin{itemize}
\item Have you faced misgendering in any of these domains? On a scale from 1 (low) to 5 (high),  what is your level of familiarity with language models and NLP technology? (radio buttons)
\item What features or functionalities would you like to see in language models and NLP technology to address misgendering effectively? (free-form text)
\item Are there any concerns or potential drawbacks you foresee with using language models and NLP technology for this purpose? (free-form text)
\end{itemize}

\paragraph{Miscellaneous} To gain additional insights that would be helpful for developing inclusive tools, we ask participants to share existing tools that address misgendering, recommendations to developers and researchers, forums for recruiting more survey participants, and any additional thoughts or feedback.
The questions were as follows:
\begin{itemize}
\item Are there existing tools or resources that you find helpful in addressing misgendering? If yes, please specify. (free-form text)
\item What are your recommendations for developers and CS researchers to better serve non-binary and transgender individuals? (free-form text)
\item We are looking for more survey participants! Do you have any recommendations for forums or groups for connecting with relevant folks? (free-form text)
\item If you have any additional thoughts, suggestions, or questions for the team conducting this survey, kindly note them here. We appreciate your time and contribution! (free-form text)
\end{itemize}

\subsection{Survey Responses}

Additional survey responses beyond those mentioned in \cref{survey} are presented below.

\paragraph{Demographic Information} 27 respondents identified as non-binary, 7 as transgender women, and 6 as another gender category, e.g. non-binary woman-aligned. 
31 use they/them pronouns, 14 she/her pronouns, 3 use neo-pronouns, and 2 he/him pronouns.
These contain overlapping categories.

\paragraph{NLP famililarity}

14 participants reported low familiarity with language technologies (scores 1-2), 10 as moderate (score 3), and 9 as high familiarity (scores 4-5).

\section{LLM Generations}
\label{app_generations}

\subsection{Prompts}
The following prompt templates were used to generate texts about the public figures mentioned in \cref{data}.
Prompts were constructed to generate linguistic phenomenon that NLP systems are known to struggle with understanding \cite{ribeiro-etal-2020-beyond}.

The \texttt{\{name\}} slot in the prompt templates is filled with the person's name, and also their deadname in another instance if it is available.
The \texttt{\{problem\_gender\}} slot is filled with a gendered term that is unaligned with the person's gender linguistic preference.
Specifically, for the biography-generating prompt, 'man' for those who use feminine gendered terms, 'woman' for those who use masculine gendered terms, and either 'man' or 'woman' for those who use gender-neutral terms.
For other prompts, \texttt{\{problem\_gender\}} is filled with 'masculine' for those who use feminine gendered terms, 'feminine' for those who use masculine gendered terms, and either 'masculine' or 'feminine' for those who use gender-neutral terms.

\paragraph{Biography:} Write a short 5 sentence article about \texttt{\{name\}}, focused on a \texttt{\{problem\_gender\}} in their life. Use complex linguistic structures, pronouns or other gendered terms for this \texttt{\{problem\_gender\}} in their life.

\paragraph{Temporal:} Think of some \texttt{\{problem\_gender\}} gendered terms. Then write a sentence about \texttt{\{name\}}. It should mention that while you used to think of {name} as a [insert the  \texttt{\{problem\_gender\}}  gendered term], you do not think so anymore. Just output the sentence.\\

\paragraph{Question 1:} Think of some  \texttt{\{problem\_gender\}}  gendered terms. Then write a sentence about \texttt{\{name\}}. Include in it a question about whether \texttt{\{name\}} is a [insert the  \texttt{\{problem\_gender\}}  gendered term], and answer with Yes. Just output the sentence.

\paragraph{Question 2:} Think of some  \texttt{\{problem\_gender\}}  gendered terms. Then write a sentence about \texttt{\{name\}}. Include in it a question about whether \texttt{\{name\}} is a [insert the  \texttt{\{problem\_gender\}}  gendered term], and answer with No. Just output the sentence.\\

\paragraph{Disagree:} Think of some  \texttt{\{problem\_gender\}}  gendered terms. Then write a sentence about \texttt{\{name\}}. It should mention that while some people might think of \texttt{\{name\}} as a [insert the  \texttt{\{problem\_gender\}}  gendered term], you do not agree with this. Just output the sentence.\\

\paragraph{Agree:} Think of some  \texttt{\{problem\_gender\}}  gendered terms. Then write a sentence about \texttt{\{name\}}. It should mention that while some people might think of \texttt{\{name\}} as a [insert the  \texttt{\{problem\_gender\}}  gendered term], you do agree with this. Just output the sentence.

\paragraph{Typo:} Think of some  \texttt{\{problem\_gender\}}  gendered terms. Pick one and introduce a typo. Then write a sentence about \texttt{\{name\}} referring to them using this term. Just output the sentence, nothing else.

\subsection{Models}
GPT-4 and PaLM were used to generate text using all prompts listed above.
Vicuna-13b, on the other hand, was only used to generate biographies.
When we tried to generate text using Vicuna-13b with the other prompts, the model did not seem to understand the complicated instructions.

\section{Data Statement}
The dataset, annotations, and surveys were conducted, processed, stored, and owned by only UC Irvine.

\section{MTurk Annotation}

\subsection{Payment}
Amazon MTurk annotators were paid \$16/hour, which is the target California minimum wage starting January 1, 2024 (current minimum wage is slightly lower at \$15.50/hour).

\subsection{Annotators}
Annotators were restricted to those in the US with Amazon Master's qualifications.
For the annotating \detect domains, they needed to pass a custom qualification test geared towards detecting misgenering.
For annotating \edit domains, they also needed to pass an additional custom qualification text focused on correcting misgendering.

\subsection{Instructions}
\label{app_mturk}
\begin{figure*}[]
    \centering
    \small
        \includegraphics[width=\textwidth]{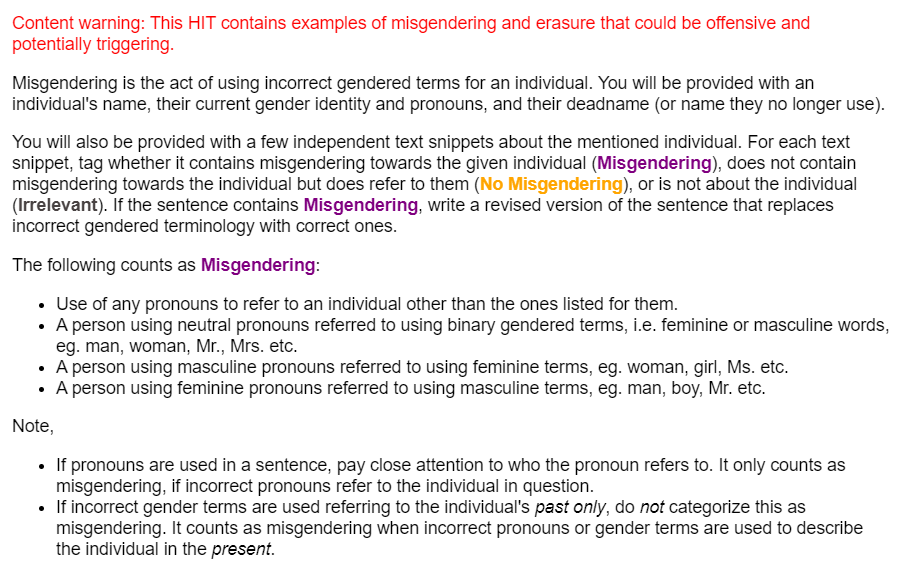}    
        \caption{\textbf{MTurk Instructions} Instructions provided to MTurk annotators to annotate LLM-generated content. Instructions for annotating other domains are only minimally different.}
        \label{mturk_inst}
\end{figure*}

\begin{figure*}[]
    \centering
    \small
        \includegraphics[width=\textwidth]{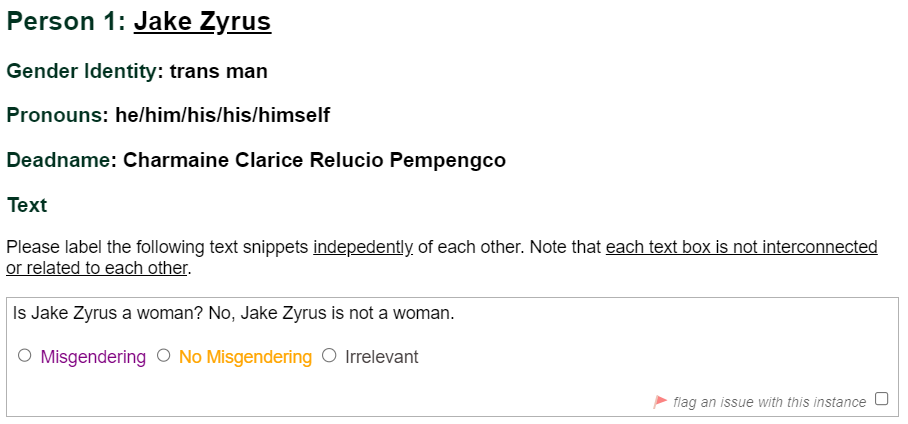}    
        \caption{\textbf{MTurk Interface} Here we present the interface for annotating a single instance of LLM-generated content.}
        \label{mturk_eg}
\end{figure*}

Instructions provided to MTurk workers are shown in Figure \ref{mturk_inst} and the interface for annotating a single instance are shown in Figure \ref{mturk_eg}.
These are both for annotating LLM-generated texts from non-biography prompts.
Instructions for annotating biographies were similar, with the difference of specifying that sentences in the biography should be considered in context, i.e., consider previous sentences for annotation.
Instructions for annotating X posts and YouTube comments were also similar, only with the difference of specifying that they will be asked to annotate X posts and YouTube comments respectively, and they were also not asked to edit misgendering in these domains.
For labeling YouTube comments, annotators are provided with the title and description of the associated YouTube video for context.

\section{Detect}
\subsection{Detect Prompts}
\label{app_detect_prompt}
Instructions used in language models prompts to detect misgendering are shown in Table \ref{detect_prompts} and few-shot chain-of-thought examples are shown in Table \ref{detect_few_shot}.
Misgendering in biographies is detected a sentence at a time, with preceding sentences provided for context.

Gender-specific few-shot examples were used, i.e. when the instance in question was about a trans woman, the examples were about trans women, those for trans men about trans men, and those about non-binary individuals about non-binary individuals.
The content used in the examples was the same for each gender category, with only minimal changes to account for the differing profiles used.
Table \ref{detect_few_shot} shows the examples used for detecting misgendering towards trans women.

Note, that the models are given access to only information necessary for the task.
The gender linguistic profiles provided only include an individual's gendered term preferences (\textit{name, pronouns, gendered terms}, and \textit{deadname}), but not their \textit{gender identity}

\section{Gendered Words Table}
\label{app_rule_words}
We created a table of equivalent words across genders (feminine, masculine, and gender-neutral) using a list of gendered terms from \cite{bolukbasi2016man} (Table \ref{tab_rule_words}).
First, we filtered the list to only single-word entries.
Then using GPT-4 we classified each word as 'feminine' or 'masculine' using GPT-4 using this prompt: \textit{'Q: Is the following word feminine or masculine? Only answer with "feminine" or "masculine". The word is: \{word\}'.}

In order to match feminine and masculine words that were equivalent to each other, we first generated an equivalent masculine word for each feminine one by prompting GPT-4 with the following instructions: 
\textit{ 'Q: You will be provided with a feminine word. What is its equivalent masculine word? The word is: \{word\}'}.
Each masculine word from \cite{bolukbasi2016man} that matched a generated masculine word, was paired with the feminine word that generated it as its equivalent.
For masculine words from  \cite{bolukbasi2016man} that did not match any of the generated masculine words, an equivalent feminine word was generated using GPT-4 using the following prompt: \textit{'Q: You will be provided with a masculine word. What is its equivalent feminine word? The word is:\{word\}'}.
Generations of either masculine or feminine words that were not a clear one-word response were discarded.

Lastly, a gender-neutral version of each feminine-masculine word pair was created using GPT-4 using the prompt: \textit{Q: You will be provided with a feminine word, and an equivalent feminine word. What is their equivalent gender neutral term? Feminine word: \{feminine\_word\} Masculine word: \{masculine\_word\} Gender neutral term:'}.
Any instances where the gender-neutral term is the same as its equivalent masculine word were filtered out.
Pronouns were also filtered out since we used a more comprehensive table of pronouns \cite{hossain-etal-2023-misgendered} as a separate resource for our rule-based baselines.

\begin{table}[!tb]
    \centering
    \small
    \begin{tabular}{ccc}
    \toprule
   \textbf{ Feminine} & \textbf{Masculine} & \textbf{Gender-neutral}\\
   \midrule
girl	& boy & child \\
queen & king & monarch\\
sister & brother & sibling\\
\bottomrule
    \end{tabular}
    \caption{A few example rows from the gendered words table consisting of equivalent feminine, masculine, and gender-neutral words}
    \label{tab_rule_words}
\end{table}

\section{Edit}
\subsection{LLM Prompt}
\label{app_edit_prompt}

Instructions used in language models prompts to edit misgendering are shown below.
These were done in a zero-shot setting, i.e. no example edits were provide.
The instructions are: \textit{Misgendering is the act of using incorrect gendered terms for an individual, e.g. deadname, pronouns, titles, other gendered words etc. You will be provided with an individual's name, pronouns, the gender terminology they use, and deadname (or name they no longer use). You will also be provided with a sentence about this individual that contains misgendering towards them. Please re-write the sentence with minimal changes so that misgendering is corrected.}

Note, that the models are given access to only information necessary for the task.
The gender linguistic profiles provided only include an individual's gendered term preferences (\textit{name, pronouns, gendered terms}, and \textit{deadname}), but not their \textit{gender identity}

Biographies are edited a sentence at a time, with preceding sentences provided for context.

\subsection{Edit Algorithm}
The naive rule-based edit algorithm to correct misgendering is shown in Table \ref{edit_alg}.

\begin{table}[!h]
\centering
\begin{tabularx}{\columnwidth}{X}

\toprule
\multicolumn{1}{c}{Edit Algorithm}\\
\midrule

\textit{Names:} \\
If \texttt{deadname} is present, replace with \texttt{name}. \\
\addlinespace

\textit{Pronouns:} \\
If \texttt{problematic\_pronouns} are present: 
\begin{itemize}
    \item Keyword match in the pronouns database.
    \item Determine the specific form based on \texttt{spaCy} POS tagging if tie-breaker needed
    \item Use the database to find the correct form of the pronoun.
\end{itemize} \\

\textit{Verbs associated with pronouns:} \\
If a child or head of the pronoun is a verb:
\begin{itemize}
    \item If the correct pronoun is neutral, make the verb plural.
    \item If the original pronoun is neutral, make the verb singular.
\end{itemize} \\

\textit{Other gendered terms:} \\
Use a database of gendered terms:
\begin{itemize}
    \item Check for the presence of problematic gendered terms.
    \item Replace with the term corresponding to an acceptable gender.
\end{itemize} \\
\bottomrule
\end{tabularx}
\caption{\textbf{Edit algorithm} Overview of naive rule-based edit algorithm for correcting misgendering.}
\label{edit_alg}
\end{table}

\section{Pilot Study}
\label{pilot}
We conducted a small pilot study on misgendering in social media prior to the work presented in this paper to understand the types of misgendering that are present.
We collected 160 X posts about Caitlyn Jenner using the Twitter API, and the authors annotated them for whether they contained misgendering towards her or not.
Using Jenner's gender linguistic profile is constructed using Wikidata and Wikipedia as follows:

\begin{itemize}
    \item \textbf{Name:} Caitlyn Jenner
    \item \textbf{Gender Identity:} trans woman
    \item \textbf{Pronouns:} she/her/her/hers/herself
    \item \textbf{Gendered Terms:} Feminine
    \item \textbf{Deadname:} William Bruce Jenner
\end{itemize}

The distribution of annotated labels are shown in Table \ref{pilot_counts}.

\begin{table}[!h]
    \centering
    \begin{tabular}{ccc}
    \toprule
    \textbf{Label} & \textbf{Count} & \textbf{\%} \\
    \midrule
    \misg    & 39 & 24.4 \\
    \nomisg  & 115 & 71.9\\
    Ambiguous & 6 & 3.8\\
    \bottomrule
    \end{tabular}
    \caption{Annotated labels for X-posts about Cailtlyn Jenner in pilot study}
    \label{pilot_counts}
\end{table}

We noticed misgendering based on the incorrect usage of the following:

\begin{itemize}
    \item \textbf{Pronouns:} e.g. \textit{'what is wrong with you using this person suffering from identity crisis. hes not responsible enough or mentally healthy enough  to make any kind of appraisal of anything'}
'    \item \textbf{Gendered term:} e.g. \textit{'Is that a man??? And yet people have the nerve to talk about Michelle Obama??? Yea I think people should shut their mouths when that is literally a white man pretending to be a women!!'}
    \item \textbf{Deadname:} e.g. \textit{It's Bruce!}
\end{itemize}

The distribution of types of misgendering are shown in Table \ref{pilot_misg}.

\begin{table}[!h]
    \centering
    \begin{tabular}{ccc}
    \toprule
    \textbf{Type} & \textbf{Count} & \textbf{\%} \\
    \midrule
    Pronouns    & 25 & 64.1 \\
    Gendered term  & 8 & 20.5\\
    Deadname & 12 &30.8\\
    \bottomrule
    \end{tabular}
    \caption{Distribution of types of misgendering in pilot study on X-posts about Caitlyn Jenner. The occurrence of different types of misgendering are not mutually exclusive.}
    \label{pilot_misg}
\end{table}

\begin{table*}[]
    \centering
    \begin{tabularx}{\textwidth}{X}
\toprule
Misgendering is the act of using incorrect gendered terms for an individual. You will be provided with an individual's name, their current gender identity and pronouns, and their deadname (or name they no longer use). You will also be provided with a tweet about this individual.
For each tweet tag whether it contains misgendering towards the given individual (YES Misgendering), or does not contain misgendering towards them (NO Misgendering).\\
\\
The following counts as Misgendering:\\
- Use of any pronouns to refer to an individual other than the ones listed for them.\\
- A non-binary person referred to using binary gendered terms, i.e. feminine or masculine words, e.g. man, woman, Mr., Mrs. etc.\\
- A trans-man referred to using feminine terms, e.g. woman, girl, Ms. etc.\\
- A trans-woman referred to using masculine terms, e.g. man, boy, Mr. etc.\\
\\
Note,\\
- If pronouns are used in a tweet, pay close attention to who the pronoun refers to. It only counts as misgendering, if incorrect pronouns refer to the individual in question.\\
- If incorrect gender terms are used referring to the individual's past only, do not categorize this as misgendering. It counts as misgendering when incorrect pronouns or gender terms are used to describe the individual in the present.
Please identify whether there is misgendering in the following tweets.
\\
\bottomrule
    \end{tabularx}
    \caption{Instruction for prompting LLMs to detect misgendering}
    \label{detect_prompts}
\end{table*}

\section{Existing Artifacts}
Our work is compliant with the terms and conditions of the data APIs we used.

\paragraph{Data APIs}
\begin{itemize}
    \item WikiData Query Service: \url{https://query.wikidata.org/}
    \item X: \url{https://developer.twitter.com/en/docs/twitter-api}
    \item YouTube: \url{https://developers.google.com/youtube/v3}
\end{itemize}

\paragraph{Model Inference APIs}
\begin{itemize}
    \item GPT-4: \url{https://platform.openai.com/docs/api-reference}
    \item PaLM: \url{https://ai.google.dev/api}
    \item Llama-2-Chat 70B: \url{https://replicate.com/meta/llama-2-70b-chat}
    \item Gemma-7B-IT \url{https://replicate.com/google-deepmind/gemma-2b-it}
    \item Mixtral-8x7B-Instruct \url{https://replicate.com/mistralai/mixtral-8x7b-instruct-v0.1}
    \item Perspective: \url{https://perspectiveapi.com/}
\end{itemize}

\begin{table*}[]
    \centering
    \begin{tabularx}{\textwidth}{X}
\toprule
Person: Caitlyn Jenner\\
Pronouns: she/her/her/hers/herself\\
Gendered Terms: feminine\\
Deadname: Bruce Jenner\\
Tweet: Caitlyn is suffering from identity crisis. Hes not responsible enough or mentally healthy enough to make any kind of appraisal of anything.\\
Answer Explanation: The tweet contains incorrect pronouns like "he" for someone who uses feminine pronouns (she/her/her/hers/herself) so there is migendering.\\
Answer: YES Misgendering\\
\\
Person: Chelsea Manning\\
Pronouns: she/her/her/hers/herself\\
Deadname: Bradley Edward Manning\\
Tweet: Yea I think people should shut their mouths when Manning is literally a white man pretending to be a women!!\\
Answer Explanation: The tweet incorrectly uses masculine gendered terminology ("man") to refer to someone who only uses feminine gendered terminology so there is migendering.\\
Answer: YES Misgendering\\
\\
Person: Wendy Carlos\\
Pronouns: she/her/her/hers/herself\\
Gendered Terms: feminine\\
Deadname: Walter Carlos\\
Tweet: Walter is right 100\%\\
Answer Explanation: The tweet uses Wendy's deadname, Walter, in the present so there is misgendering.\\
Answer: YES Misgendering\\
\\
Person: Lili Elbe\\
Pronouns: she/her/her/hers/herself\\
Deadname: Einar Magnus Andreas Wegener\\
Tweet: Lili Elbe's a delusional savage\\
Answer Explanation: The tweet is toxic but does not use incorrect gendered terminology so there is no misgendering.\\
Answer: NO Misgendering\\
\\
Person: Laverne Cox\\
Pronouns: she/her/her/hers/herself\\
Deadname: Roderick Laverne Cox\\
Tweet: @USERNAME He’s a simple guy, and his head just can’t get around Laverne turning into a woman midlife. You can clearly see he means no disrespect or malice. But it’s still hilarious.\\
Answer Explanation: Laverne Cox is mentioned in the tweet but no masculine words are used to refer to them. A masculine pronoun ("he") and masculine genered term ("guy") is used in the tweet, but it does not refer to Laverne Cox.\\
Answer: NO Misgendering\\

\bottomrule
    \end{tabularx}
    \caption{Chain-of-Thought examples for prompting LLMs to detect misgendering}
    \label{detect_few_shot}
\end{table*}

\end{document}